\newcommand{\abs}[1]{\left\vert#1\right\vert}

\newcommand{\Real}{\mathbb R}

\newcommand{\Eq}[1]{\begin{equation}#1\end{equation}}
\newcommand{\Vc}[1]{\mbox{\boldmath$#1$}}

\newcommand{\vx}{\Vc{x}}
\newcommand{\vX}{\Vc{X}}

\newcommand{\vY}{\Vc{Y}}

\newcommand{\vv}{\Vc{v}}
\newcommand{\vu}{\Vc{u}}

\newcommand{\mX}{\mathcal{X}}

\newcommand{\mM}{\mathcal{M}}
\newcommand{\mI}{\mathcal{I}}
\newcommand{\mN}{\mathcal{N}}

\newcommand{\mS}{\mathcal{S}}
\newcommand{\mP}{\mathcal{P}}

\DeclareMathAlphabet{\mathsfsl}{OT1}{cmss}{m}{sl}

\documentclass[11pt,journal,draftcls,onecolumn]{IEEEtran}

\usepackage{setspace,pifont,calc,amsmath,amsfonts,amsbsy,amssymb,tabularx,float,times}
\usepackage{cite,url,graphicx,subfigure,epsfig,algorithm,algorithmic,multirow}

\begin{document}

\title{FINE: Fisher Information Non-parametric Embedding}%
\author{Kevin M. Carter$^1$\thanks{{\bf Acknowledgement}: This work is
partially funded by the National Science Foundation, grant No.
CCR-0325571.}, Raviv Raich$^2$, William G. Finn$^3$, and Alfred O. Hero III$^1$\\
$^{1}$ Department of EECS,    University of Michigan,    Ann Arbor,
MI 48109\\
$^{2}$ School of EECS, Oregon State University, Corvallis, OR 97331\\
$^{3}$ Department of Pathology,    University of Michigan,    Ann Arbor,
MI 48109\\
{\normalsize \tt  \{kmcarter,wgfinn,hero\}@umich.edu}, {\normalsize \tt
raich@eecs.oregonstate.edu}}
\maketitle

\begin{abstract}
We consider the problems of clustering, classification, and visualization of high-dimensional data when no straightforward Euclidean representation exists. Typically, these tasks are performed by first reducing the high-dimensional data to some lower dimensional Euclidean space, as many manifold learning methods have been developed for this task. In many practical problems however, the assumption of a Euclidean manifold cannot be justified. In these cases, a more appropriate assumption would be that the data lies on a \emph{statistical} manifold, or a manifold of probability density functions (PDFs). In this paper we propose using the properties of information geometry in order to define similarities between data sets using the Fisher information metric. We will show this metric can be approximated using entirely non-parametric methods, as the parameterization of the manifold is generally unknown. Furthermore, by using multi-dimensional scaling methods, we are able to embed the corresponding PDFs into a low-dimensional Euclidean space. This not only allows for classification of the data, but also visualization of the manifold. As a whole, we refer to our framework as Fisher Information Non-parametric Embedding (FINE), and illustrate its uses on a variety of practical problems, including bio-medical applications and document classification.
\end{abstract}

\section{Introduction}
The fields of statistical learning and machine learning are used to study problems of inference, which is to say gaining knowledge through the construction of models in order to make decisions or predictions based on observed data \cite{Bousquet:ALML04}. Statistical learning examines problems such as observing natural associations between data sets (clustering), and predicting to which class of known groupings an unlabeled data set belongs (classification), based on some model defined by a priori knowledge of the data. Machine learning introduces a non-parametric approach to these learning tasks via model-free learning from examples. Recent work on manifold learning aims at the high dimension regime, in which examples are governed by geometrical constraints effectively reducing the dimension of the problem from a high extrinsic dimension to a low intrinsic dimension. On the other hand, information geometry aims at understanding the structure of statistical models and introduces a geometric perspective to inference problems \cite{Amari&Nagaoka:2000}.

We are interested in the cross section of the three fields; using the principles of each to solve problems that do not fit within the framework of any of the individual fields. Often data does not exhibit a low intrinsic dimension in the data domain as one would have in manifold learning. A straightforward strategy is to express the data in terms of a low-dimensional feature vector for which the \emph{curse of dimensionality} is alleviated. This initial processing of data as real-valued feature vectors in Euclidean space, which is often carried out in an ad hoc manner, has been called the "dirty laundry" of machine learning \cite{Dietterich:2002}. This procedure is highly dependent on having a good model for the data and in the absence of such model may be highly suboptimal. When a statistical model is available, the process of obtaining a feature vector can be done optimally by extracting the model parameters for a given data set and thus characterizing the data through its lower dimensional parameter vector. We are interested in extending this approach to the case in which the data follows an unknown parametric statistical model. 

While the problem of learning in a Euclidean space is well defined, there are many problems in which the data cannot be appropriately represented by a Euclidean manifold, and the model parameters are unspecified and must be learned through the data. In flow cytometry, pathologists study blood samples containing many cells taken from a patient. Each individual cell is analyzed with different fluorescent markers, resulting in a large, high-dimensional data set. This is assumed to be a realization of some overriding parametric model, but the model parameters are unknown. Pathologists desire the ability to appropriately classify patients with differing ailments that may express similar responses to these markers. For the purposes of analysis and visualization, it is then necessary to reduce the dimensionality of these sets. The problem of document classification is one in which the data is clearly non-Euclidean, as each set is a collection of words from a dictionary. It is still desired to distinguish between documents by forming clusters of different similarities. A standard method is to form a probability distribution over a dictionary and use methods of information geometry to determine a similarity between data sets \cite{Lebanon:ISIGA05}. Applications of statistical manifolds have also been presented in the cases of face recognition \cite{Arandjelovic:CVPR05}, texture segmentation \cite{Lee:ICIP05}, image analysis \cite{Srivastava:CVPR07}, and shape analysis \cite{Kim:LIDS05}.

A common theme to all of the problems presented above is that the model from which the data is generated is unknown. In this paper, we present a framework to handle such problems. Specifically, we focus on the case where the data is high-dimensional and no lower dimensional Euclidean manifold gives a sufficient description. In many of these cases, a lower dimensional statistical manifold can be used to assess the data for various learning tasks. We refer to our framework as Fisher Information Non-parametric Embedding (FINE), and it includes characterization of data sets in terms of a non-parametric statistical model, a geodesic approximation of the Fisher information distance as a metric for evaluating similarities between data sets, and a dimensionality reduction procedure to obtain a low-dimensional Euclidean embedding of the original high-dimensional data set for the purposes of both classification and visualization.

Statistical manifolds in both the parametric and non-parametric settings have been well discussed \cite{Pistone&Rogantin:B99,Cena:Thesis02}. Our work differs in that we assume the manifold is derived from some natural parameterization, only that set of parameters is unknown. There has been much work presented on the use of statistical manifolds \cite{Lebanon:TIT05,Lebanon:ISIGA05,Lafferty&Lebanon:TIT05,Srivastava:CVPR07} and information geometry \cite{Salojarvi:ICANN03,Yeang:ECML02} in learning problems, all proposing alternatives to using Euclidean geometry for data modeling. These methods focus on clustering and classification, and do not explicitly address the problems of dimensionality reduction (embedding each set into a low-dimensional Euclidean space) and visualization. Additionally, they focus on parameter estimation as a necessity for their methods, as opposed to our work which is performed in a non-parametric setting. We provide a start-to-finish framework which enables analysis of high-dimensional data through non-linear embedding into a low-dimensional space by information, not Euclidean, geometry. Our methods require no explicit model assumptions; only than that the given data is a realization from an unknown model with some natural parameterization.

Recent work by Lee \emph{et al}.~\cite{Lee&Abbott:CVPR07} similar to our own \cite{Carter&Hero:Allerton07, Carter&Raich:ICASSP08} has demonstrated the use of statistical manifolds for dimensionality reduction. While each work has been developed independently and originally presented at nearly the same time, they share enough similarities that we now express the different contributions of our own work. Specifically, we consider the work presented by Lee \emph{et al}.~to be a specialized case of our more general framework. They focus on the specific case of image segmentation, which consists of multinomial distributions as points which lie on an $n$-simplex (or projected onto an $n+1$-dimensional sphere). By framing their problem as such, they are able to exploit the properties of such a manifold: using the cosine distance as an exact computation of the Fisher information distance, and using linear methods (PCA) of dimensionality reduction. They have shown very promising results for the problem of image segmentation, and briefly mention the possibility of using non-linear methods of dimensionality reduction, which they consider unnecessary for their problem. The work we present differs in that we make no assumptions on the type of distributions making up the statistical manifold. As such, our geodesic approximation for the Fisher information accounts for submanifolds of interest. This is illustrated later in Fig.~\ref{f:submanifold}, where the submanifold lies on the $n+1$-dimensional sphere, but does not fill the entire space. As such, there is no exact measure of the Fisher information between points, and we must approximate with a geodesic along the manifold. Additionally, we utilize non-linear methods of dimensionality reduction, which we consider to be more relevant for many non-linear types of applications. Finally, by considering all statistical manifolds rather than focusing on those of consisting of multinomial distributions, we are able to apply our methods to many problems of practical interest.

This paper is organized as follows: Section \ref{S:Background} describes a background in information geometry and statistical manifolds. Section \ref{S:ProblemFormulation} gives the formulation for the problem we wish to solve, while Section \ref{S:Techniques} develops and outlines the FINE algorithm. We illustrate the results of using FINE on real and synthetic data sets in Section \ref{S:Apps}. Finally, we draw conclusions and discuss the possibilities for future work in Section \ref{S:Conclusions}.
\section{Background on Information Geometry}
\label{S:Background}
Information geometry is a field that has emerged from the study of geometrical structures on manifolds of probability distributions. These investigations analyze probability distributions as geometrical structures in a Riemannian space. Using tools and methods deriving from differential geometry, information geometry is applicable to information theory, probability theory, and statistics. The field of information theory is largely based on the works of Shun'ichi Amari \cite{Amari:90} and has been used for analysis in such fields as statistical inference, neural networks, and control systems. In this section, we will give a brief background on the methods of information geometry that we utilize in our framework. For a more thorough introduction to information geometry, we suggest \cite{Kass&Vos:97} and \cite{Amari&Nagaoka:2000}.
\subsection{Differential Manifolds}
\label{SS:DiffMan}
The concept of a differential manifold is similar to that of a smooth curve or surface lying in a high-dimensional space. A manifold $\mM$ can be intuitively thought of as a set of points with a coordinate system. These points can be from a variety of constructs, such as Euclidean coordinates, linear system, images, or probability distributions. Regardless of the definition of the points in the manifold $\mM$, there exists a coordinate system with a one-to-one mapping from $\mM$ to $\Real^d$, and as such, $d$ is known as the dimension of $\mM$.

For reference, we will refer to the coordinate system on $\mM$ as $\psi:\mM\rightarrow \Real^d$. If $\psi$ has $\mM$ as its domain, we call it a global coordinate system \cite{Amari&Nagaoka:2000}. In this situation, $\psi$ is a one-to-one mapping onto $\Real^d$ for all points in $\mM$. A manifold is differentiable if the coordinate system mapping $\psi$ is differentiable over its entire domain. If $\psi$ is infinitely differentiable, the manifold is said to be `smooth' \cite{Kass&Vos:97}.

In many cases there does not exist a global coordinate system. Examples of such manifolds include the surface of a sphere, the ``swiss roll'', and the torus. For these manifolds, there are only local coordinate systems. Intuitively, a local coordinate system acts as a global coordinate system for a local neighborhood of the manifold, and there may be many local coordinate systems for a particular manifold. Fortunately, since a local coordinate system contains the same properties as a global coordinate system (only on a local level), analysis is consistent between the two. As such, we shall focus solely on manifolds with a global coordinate system.
\subsubsection{Statistical Manifolds}
\label{SS:StatMan}
Let us now present the notion statistical manifolds, or a set $\mM$ whose elements are probability distributions. A probability distribution function (PDF) on a set $\mX$ is defined as a function $p:\mX\rightarrow\Real$ in which
\Eq{ \label{Equation:PDF}
p(x)\ge0, \,\forall x\in\mX
}
\[\int p(x)\,dx=1.\]
We describe only the case for continuum on the set $\mX$, however if $\mX$ was discrete valued, equation (\ref{Equation:PDF}) will still apply by switching $\int p(x)\,dx=1$ with $\sum p(x)=1$. If we consider $\mM$ to be a family of PDFs on the set $\mX$, in which each element of $\mM$ is a PDF which can be parameterized by $\theta=\left[\theta^1,\ldots,\theta^n\right]$, then $\mM$ is known as a statistical model on $\mX$. Specifically, let
\Eq{ \label{E:statman}
\mM=\{p(x\mid\theta)\mid\theta\in\Theta\subseteq\Real^d\}
,}
with $p(x\mid\theta)$ satisfying the equations in (\ref{Equation:PDF}). Additionally, there exists a one-to-one mapping between $\theta$ and $p(x\mid\theta)$.

Given certain properties of the parameterization of $\mM$, such as differentiability and $C^\infty$ diffeomorphism (details of which are described in \cite{Amari&Nagaoka:2000}), the parameterization $\theta$ is also a coordinate system of $\mM$. In this case, $\mM$ is known as a statistical manifold. In the rest of this paper, we will use the terms `manifold' and `statistical manifold' interchangeably.
\subsection{Distances on Manifolds}
\label{SS:DistMan}
In Euclidean space, the distance between two points is defined as the length of a straight line between the points. On a manifold, however, one can measure distance by a trace of the shortest path between the points along the manifold. This path is called a geodesic, and the length of the path is the geodesic distance. In information geometry, the distance between two points on a manifold is analogous to the difference in information between them, and is defined by the Fisher information metric.
\subsubsection{Fisher Information Metric}
\label{SS:FIM}
The Fisher information measures the amount of information a random variable $X$ contains in reference to an unknown parameter $\theta$. For the single parameter case it is defined as
\[
\mI(\theta)=E\left[\left(\frac{\partial}{\partial\theta}\log f(X;\theta)\right)^2|\theta\right]
.\]
If the condition $\int\frac{\partial^2}{\partial\theta^2}f(X;\theta)\,dX=0$ is met, then the above equation can be written as
\[
\mI(\theta)=-E\left[\frac{\partial^2}{\partial\theta^2}\log f(X;\theta)\right]
.\]
For the case of multiple parameters $\theta=\left[\theta^1,\ldots,\theta^n\right]$, we define the Fisher information matrix $[\mI(\theta)]$, whose elements consist of the Fisher information with respect to specified parameters, as
\Eq{ \label{Equation:FIM}
\mI_{ij}=\int{f(X;\theta)\frac{\partial \log f(X;\theta)}{\partial\theta^i}\frac{\partial \log f(X;\theta)}{\partial\theta^j}\,dX}
.}

For a parametric family of probability distributions, it is possible to define a Riemannian metric using the Fisher information matrix, known as the information metric. The information metric distance, or Fisher information distance, between two distributions $p(x;\theta_1)$ and $p(x;\theta_2)$ in a single parameter family is
\Eq{ \label{Eq:FID_single}
    D_F(\theta_1,\theta_2)=\int_{\theta_1}^{\theta_2}{\mI(\theta)^{1/2}d\theta}
    ,}
where $\theta_1$ and $\theta_2$ are parameter values corresponding to the two PDFs and $\mI(\theta)$ is the Fisher information for the parameter $\theta$. Extending to the multi-parameter case, we obtain:
\Eq{ \label{Eq:FID_multi}
    D_F(\theta_1,\theta_2)=\min_{\theta:\theta(0)=\theta_1,\theta(1)=\theta_2} \int_{0}^{1}{\sqrt{\left(\frac{d\theta}{d\beta}\right)^T \mI(\theta) \left(\frac{d\theta}{d\beta}\right)} d\beta}
    .}

\subsubsection{Example}
\label{SS:example}
Here we present a derivation of a geodesic distance between univariate Gaussian densities via the Fisher information metric for two reasons. First, we would like to illustrate how involved the process is for such a simple family of PDFs. Secondly, we present a process of deriving the Fisher information metric that is involved in computing the geodesic distance. Let us consider the family of univariate Gaussian distributions $\mP=\{p_1,\ldots,p_n\}$, where \[p_i(x)=\frac{1}{\sqrt{2\pi\sigma_i^2}}\exp{\left(-(x-\mu_i)^2/2\sigma_i^2\right)}.\] 

For the case of $\mP$ parameterized by $\theta=\left(\frac{\mu}{\sqrt2}, \sigma\right)$, the resultant Fisher information matrix is
\[[\mI(\theta)]=\left(
    \begin{array}{cc}
      \frac{2}{\sigma^2} & 0 \\
      0 & \frac{2}{\sigma^2} \\
    \end{array}
  \right)
.\]
We omit the derivation, which can be found in \cite{Kass&Vos:97} and is straight forward from (\ref{Equation:FIM}).

We define the distance between two points on the manifold as the minimum length between all paths connecting the two points. Using the inner product associated with the Fisher information matrix
\[<\vu,\vv>_F=\vu^t[\mI(\theta)]\vv,\]
we define the length of the path $P$ between two points parameterized by $\theta_1$ and $\theta_2$, on the manifold $\mM$ as
\[
\|\theta_1-\theta_2\|_P=\sqrt{<\theta_1-\theta_2,\theta_1-\theta_2>_F}
.\]
Using the parameterization $\theta(t)$ such that $\theta(0)=\theta_1$ and $\theta(1)=\theta_2$, we obtain the length of $P$ as
\[\|\theta_1-\theta_2\|_{P}=\int_0^1{\sqrt{\left(\frac{d}{dt}\theta(t)\right)^T \mI(\theta(t)) \left(\frac{d}{dt}\theta(t)\right)}\,dt}.\]
We are able to define the distance between points $p_1=p(x;\theta_1)$ and $p_2=p(x;\theta_2)$ as the minimum over all path lengths defined above
\Eq{ \label{Equation:DF_poincare}
D_F(p_1,p_2)=\min_{\theta(t)}\sqrt2\int_0^1{\sqrt{\frac{\frac{1}{\sqrt2}\dot{\mu}^2+\dot{\sigma}^2}{\sigma(t)^2}}\,dt}
,}
where $\dot{\mu}=\frac{d}{dt}\mu(t)$ and $\dot{\sigma}=\frac{d}{dt}\sigma(t)$.

The solution to (\ref{Equation:DF_poincare}) is the well known Poincar$\rm{\acute{e}}$ hyperbolic distance, in which the shortest path between two points is the length of an arc on a circle in which both points are at a radius length from the circle's center. In the case of the univariate normal distribution, this arc is a straight line when the mean is held constant and the variance is changed.

By changing variables and parameterizing $\sigma$ as a function of $\mu$, we obtain:
\[
\min_{\sigma(\mu): \sigma(\mu_1)=\sigma_1, \sigma(\mu_2)=\sigma_2}\int_{\mu_1}^{\mu_2}\sqrt{\frac{1+\dot{\sigma}^2}{\sigma(\mu)^2}}\,d\mu
,\]
where $\dot{\sigma}=\frac{d}{d\mu}\sigma(\mu)$. It should be clear that this is a representation of (\ref{Eq:FID_single}). It should also be noted that there exists a one-to-one mapping $\sigma(\mu):\Real \rightarrow \Real^+$ along the geodesic from $\sigma(\mu_1)$ to $\sigma(\mu_2)$, except for the case when $\mu_1=\mu_2$.

Solving (\ref{Equation:DF_poincare}) becomes a problem of calculus of variations. For the univariate normal family of distributions, this has been calculated in a closed-form expression presented in \cite{Costa:ITW05}, determining the Fisher information distance as:
\Eq{ \label{Equation:FID_normal} D_F(p_1,p_2)=\sqrt{2}\log{\frac{\left\|\left(\frac{\mu_1}{\sqrt{2}},\sigma_1\right) - \left(\frac{\mu_2}{\sqrt{2}},-\sigma_2\right)\right\| + \left\|\left(\frac{\mu_1}{\sqrt{2}},\sigma_1\right) - \left(\frac{\mu_2}{\sqrt{2}},\sigma_2\right)\right\|} {\left\|\left(\frac{\mu_1}{\sqrt{2}},\sigma_1\right) - \left(\frac{\mu_2}{\sqrt{2}},-\sigma_2\right)\right\| - \left\|\left(\frac{\mu_1}{\sqrt{2}},\sigma_1\right) - \left(\frac{\mu_2}{\sqrt{2}},\sigma_2\right)\right\|}}
.}

\begin{figure}[t]
  \centerline{
  \subfigure[]{ \label{f:kl_fi_a}
    \includegraphics[scale=.55]{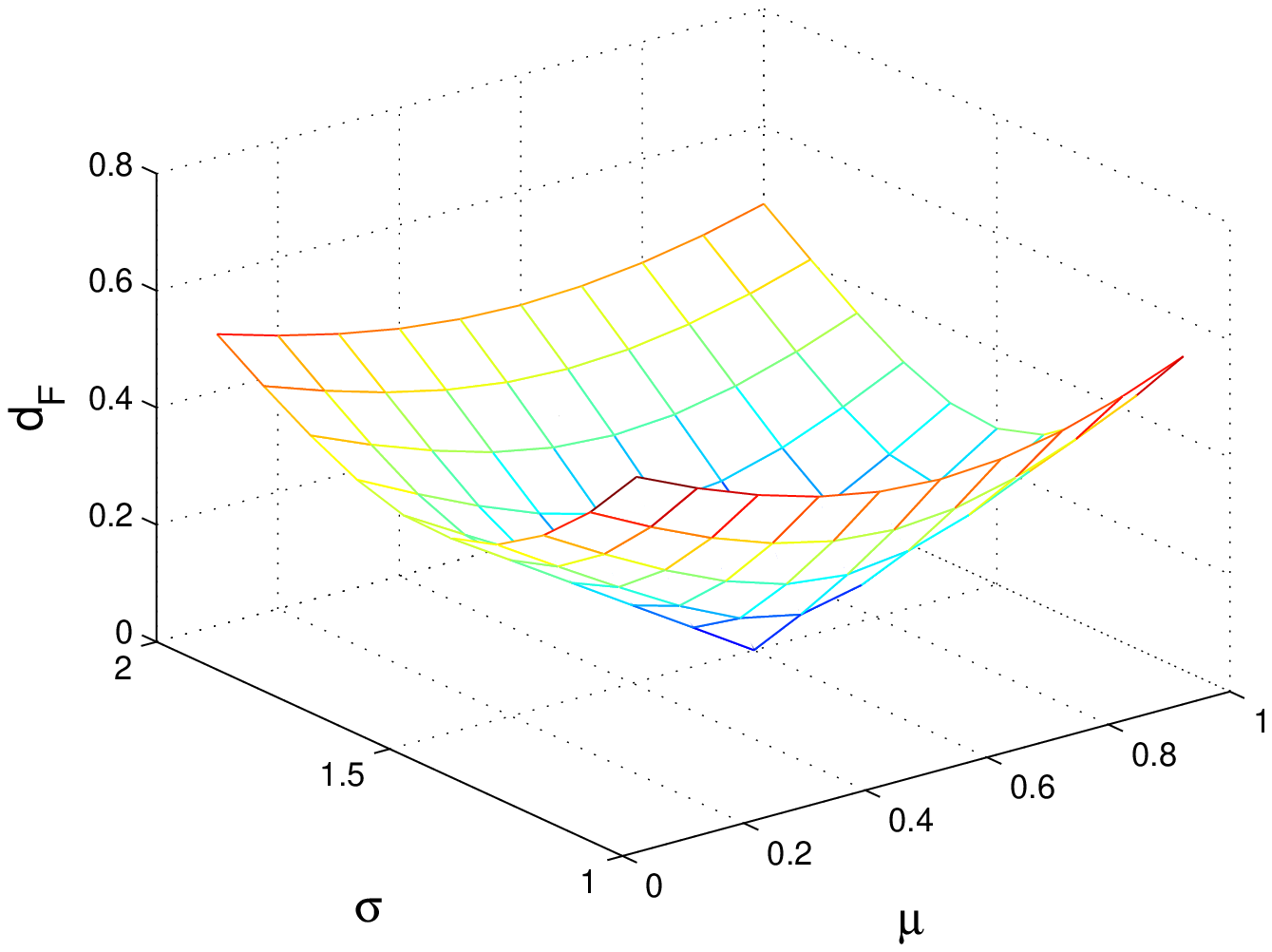}}\\
  \subfigure[]{ \label{f:kl_fi_b}
    \includegraphics[scale=.55]{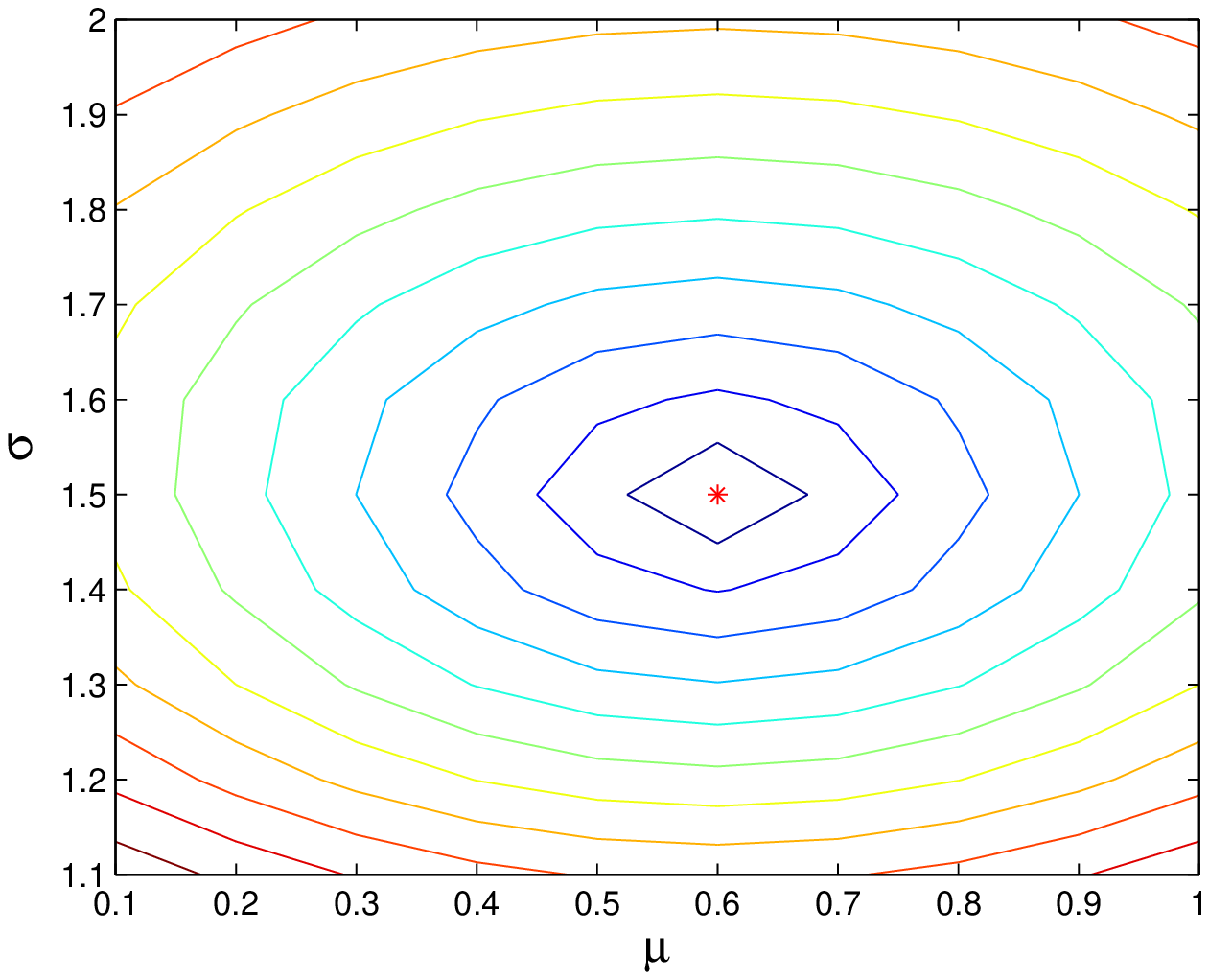}}\\
  }
  \caption{a) Mesh-grid and b) Contour plots of the Fisher information distance based on a grid of univariate normal densities, parameterized by $\left(\mu,\sigma\right)$. The reference point, $p_i$, is located at $(\mu_i,\sigma_i)=(0.6,1.5)$ and is denoted by the red star.}
  \label{f:kl_fi}
\end{figure}

For visualization, let us define a set of probability densities $\mP=\left\{p_i(x)\right\}$ on a grid, such that $p_i=p_{k,l}$ is parameterized by $\left(\mu_i,\sigma_i\right)=\left(\alpha k,1+\beta l\right)$, $k,l=1\ldots n$ and $\alpha,\beta\in \Real$. Figure \ref{f:kl_fi} shows a mesh-grid and contour plot of the Fisher information distance between the density defined by $(\mu_i,\sigma_i)=(0.6,1.5)$ and the neighboring densities on the set $\mP$ ($\alpha=\beta=0.1$).

\section{Problem Formulation}
\label{S:ProblemFormulation}
A key property of the Fisher information metric is that it is independent of the parameterization of the manifold \cite{Kass&Vos:97,Srivastava:CVPR07}. Although the evaluation remains equivalent, calculating the FIM requires knowledge of the parameterization, which is generally not available. We instead assume that the collection of density functions lie on a manifold that can be described by some natural parameterization. Specifically, we are given $\mP=\left\{p_1,\ldots,p_n\right\}$, where $p_i\in \mM$ is a PDF and $\mM$ is a manifold embedded in $\mS$, the simplex of densities in $L_1$. Under these circumstances, it is important to note that much of the same theory still applies for determining dissimilarity between probability distributions. Our goal is to find an approximation for the geodesic distance between points on $\mM$ using only the information available in $\mP$. Can we find an approximation function $G$ which yields
\Eq{ \label{E:ProbForm}
\hat{D}_F(p_i,p_j)=G(p_i,p_j;\mP)
,}
such that $\hat{D}_F(p_i,p_j)\rightarrow D_F(p_i,p_j)$ as $n\rightarrow\infty$?

This problem is similar to the setting of classical papers \cite{Tenenbaum&etal:Science00,Belkin&Niyogi:NIPS02} in manifold learning and dimensionality reduction, where only a set of points on the manifold are available. As such, we are able to use these manifold learning techniques to construct a low-dimensional embedding of that family. This not only allows for an effective visualization of the manifold (in 2 or 3 dimensions), but by reducing the effect of the \emph{curse of dimensionality} we can perform clustering and classification on the family of distributions lying on the manifold.

\subsection{Approximation of Fisher Information Distance}
\label{SS:ApproxFish}
The Fisher information distance is consistent, regardless of the parameterization of the manifold \cite{Srivastava:CVPR07}. This fact enables the approximation of the information distance when the specific parameterization of the manifold is unknown, and there have been many metrics developed for this approximation. An important class of such divergences is known as the $f$-divergence \cite{Csiszar:67}, in which $f(u)$ is a convex function on $u > 0$ and
\[
    D_f(p\|q)=\int{p(x)f\left(\frac{q(x)}{p(x)}\right)}
    .\]

A specific and important example of the $f$-divergence is the $\alpha$-divergence, where $D^{(\alpha)}=D_{f^{(\alpha)}}$ for a real number $\alpha$. The function $f^{(\alpha)}(u)$ is defined as
\[
        f^{(\alpha)}(u)=\left\{
                          \begin{array}{cl}
                            \frac{4}{1-\alpha^2}\left(1-u^{(1+\alpha)/2}\right) & \alpha\neq\pm1 \\
                            u\log u & \alpha=1 \\
                            -\log u & \alpha=-1
                          \end{array}
                        \right.
.\]

As such, the $\alpha$-divergence can be evaluated as
\[
    D^{(\alpha)}(p\|q)=\frac{4}{1-\alpha^2}\left(1-\int{p(x)^\frac{1-\alpha}{2}q(x)^\frac{1+\alpha}{2}dx}\right)\quad\alpha\neq1
,\]
and
\Eq{ \label{E:Dalpha-1}
    D^{(-1)}(p\|q)=D^{(1)}(q\|p)=\int{p(x)\log \frac{p(x)}{q(x)}}
.}
The $\alpha$-divergence is the basis for many important and well known divergence metrics, such as the Hellinger distance, the Kullback-Leibler divergence, and the Renyi-Alpha entropy \cite{Renyi:MSP61}.
%

\subsubsection{Kullback-Leibler Divergence}
\label{SS:KLD}
The Kullback-Leibler (KL) divergence is defined as
\Eq{\label{Equation:KL}
KL(p\|q)=\int{p(x)\log \frac{p(x)}{q(x)}}
 ,}
which is equal to $D^{(-1)}$ (\ref{E:Dalpha-1}). The KL-divergence is a very important metric in information theory, and is commonly referred to as the relative entropy of one PDF to another. Kass and Vos show \cite{Kass&Vos:97} the relation between the Kullback-Leibler divergence and the Fisher information distance is
\[\sqrt{2KL(p\|q)}\rightarrow D_{F}(p,q)\]
as $p\rightarrow q$. This allows for an approximation of the Fisher information distance, through the use of the available PDFs, without the need for the specific parameterization of the manifold.

Returning to our illustration developed in Section \ref{SS:example}, we have defined the data set $\mP$ of univariate normal distributions, and presented an expression for the Fisher information distance on the resultant manifold (\ref{Equation:FID_normal}). The Kullback-Leibler divergence between univariate normal distributions is also available in a closed-form expression:
\[KL(p_i\|p_j)=\frac{1}{2}\left(\log\left(\frac{\sigma_j^2}{\sigma_i^2}\right) + \frac{\sigma_i^2}{\sigma_j^2} + \left(\mu_j-\mu_i\right)^2 / \sigma_j^2 - 1\right)
.\]

\begin{figure}[t]
  \centerline{
    \subfigure[Mesh-grid]{\label{error:sub:a}
      \includegraphics[scale=.55]{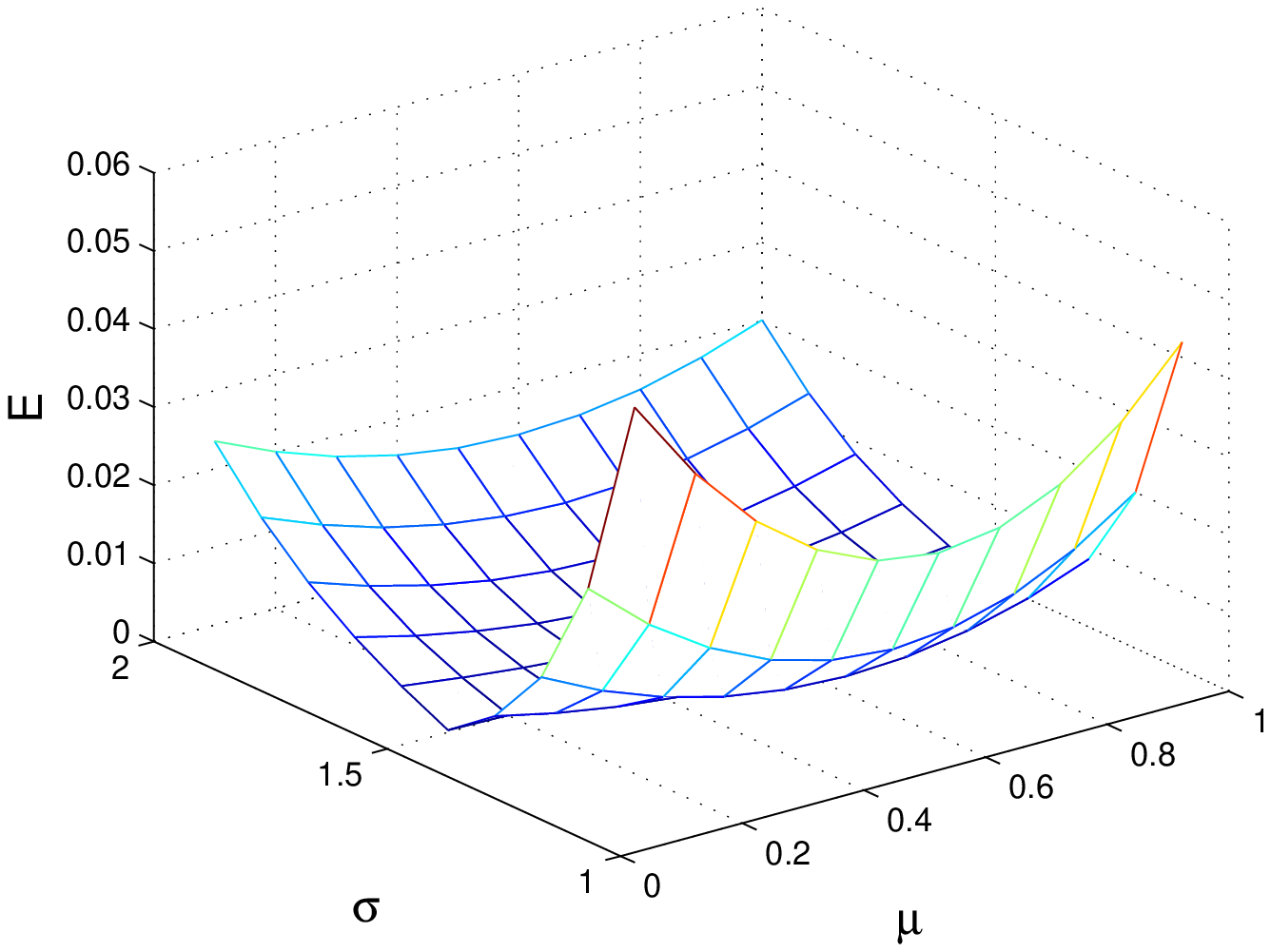}
    }
    \subfigure[Contour plot]{\label{error:sub:b}
      \includegraphics[scale=.55]{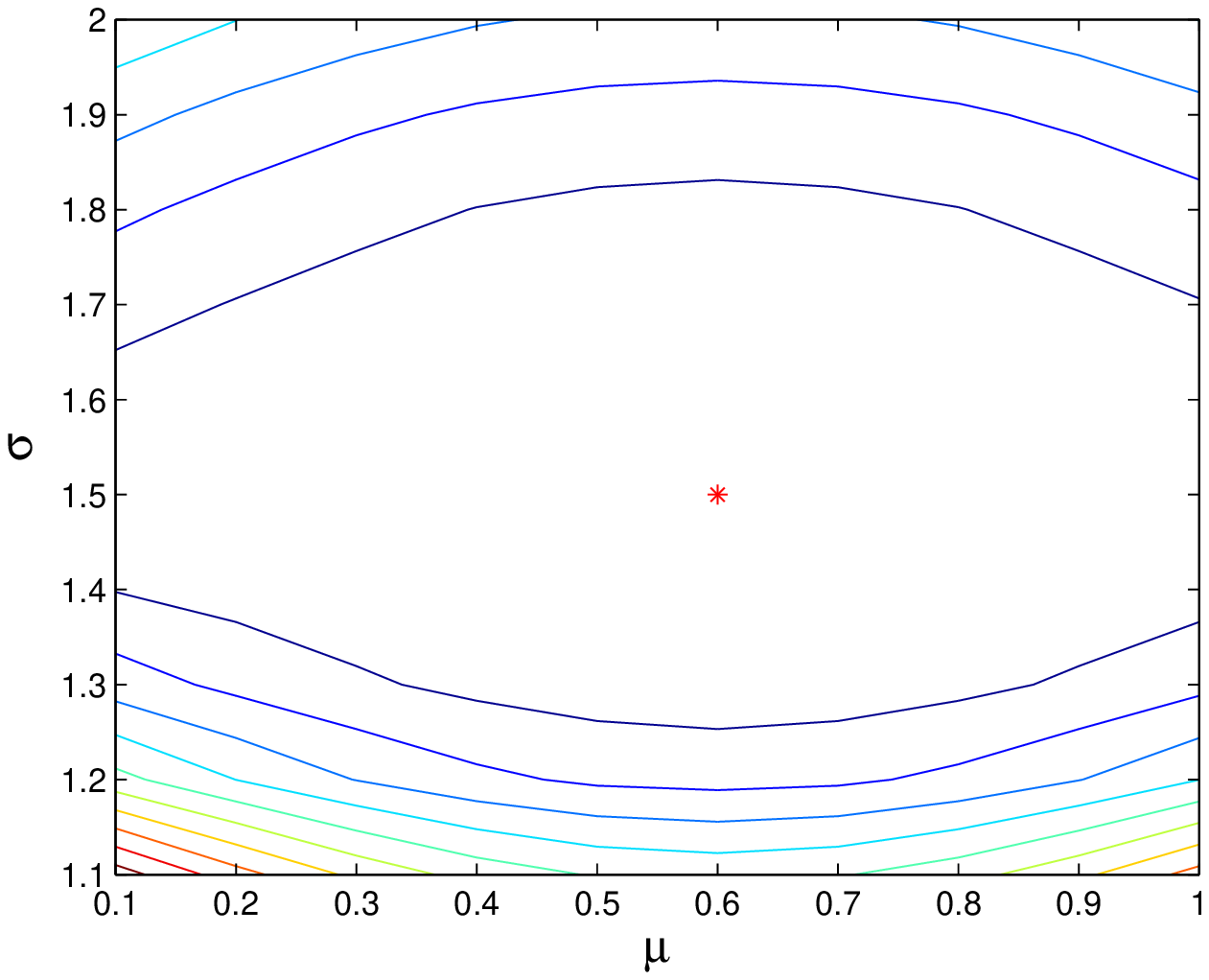}
    }
  }
  \caption{a) Mesh-grid and b) Contour plots of the error between the KL-divergence and the Fisher information distance based on a grid of univariate normal densities, parameterized by $\left(\mu,\sigma\right)$. $E=\abs{\sqrt{2KL(p_i\|p_j)}-D_{F}(p_i,p_j)}$. Note that $\sqrt{2KL}\rightarrow D_F$, where $p_i$ is denoted by the red star.}
  \label{f:kl_fi_error}
\end{figure}

To compare the KL-divergence to the Fisher information distance, we define the error as $E=\abs{\sqrt{2KL(p_i\|p_j)}-D_{F}(p_i,p_j)}$, where $p_{i,j}\in\mP$. In Fig.~\ref{f:kl_fi_error} we display the mesh-grid and contour plots of $E$, where point $p_i$ is held constant in the center of the grid defining $\mP$, and $p_j$ varies about the manifold. As described earlier, as the density $p_j\rightarrow p_i$, the error $E\rightarrow 0$. In Fig.~\ref{f:kl_fi_error}(b), the reference point $p_i$ is noted by the red star.

It should be noted that the KL-divergence is not a distance metric, as it does not satisfy the symmetry, $KL(p\|q)\neq KL(p\|q)$, or triangle inequality properties of a distance metric. To obtain this symmetry, we will define the KL-divergence as:
\Eq{ \label{Equation:DKL}
        D_{KL}(p,q) = KL(p\|q)+KL(q\|p)
    ,}
which is symmetric, but still not a distance as it does not satisfy the triangle inequality. Since the Fisher information is a symmetric measure, we can relate the symmetric KL-divergence
and approximate the Fisher information distance as
\Eq{ \label{E:FisherApprox}
\sqrt{D_{KL}(p,q)}\rightarrow D_F(p,q)
,}
as $p\rightarrow q$.

\subsubsection{Hellinger Distance}
\label{SS:Hellinger}
Another important result of the $\alpha$-divergence is the evaluation with $\alpha=0$:
\[
    D^{(0)}(p\|q)=2\int{\left(\sqrt{p(x)}-\sqrt{q(x)}\right)^2dx},
\]
which is called the closely related to the Hellinger distance,
\[
    D_H=\sqrt{\frac{1}{2}D^{(0)}}
,\]
which satisfies the axioms of distance - symmetry and the triangle inequality.  The Hellinger distance is related to the information distance in the limit by
\[2D_{H}(p,q)\rightarrow D_{F}(p,q)
\]
as $p\rightarrow q$ \cite{Kass&Vos:97}. We note that the Hellinger distance is related to the Kullback-Leibler divergence, as in the limit $\sqrt{KL(p\|q)}\rightarrow D_H(p,q)$.

\subsubsection{Other Fisher Approximations}
\label{SS:Others}
There are other metrics which approximate the Fisher information distance, such as the cosine distance. When dealing with multinomial distributions, the approximation
\[D_C(p,q)=2\arccos \int{\sqrt{p\cdot q}} \to D_F(p,q),\]
is the natural metric on the sphere.

We restrict our analysis to that of the Kullback-Leibler divergence and the Hellinger distance. The KL-divergence is a great means of differentiating shapes of continuous PDFs. Analysis of (\ref{Equation:KL}) shows that as $p(x)/q(x)\rightarrow \infty$, $KL(p\|q)\rightarrow \infty$. These properties ensure that the KL-divergence will be amplified in regions where there is a significant difference in the probability distributions. This cannot be used in the case of a multinomial PDF, however, because of divide-by-zero issues. In that case the Hellinger distance is the desired metric as there exists a monotonic transformation function $\psi:D_H\to D_C$ \cite{Kass&Vos:97}. For additional measures of probabilistic distance, some of which approximate the Fisher information distance, and a means of calculating them between data sets, we refer the reader to \cite{Zhou&Chellappa:PAMN06}.


\subsection{Approximation of Distance on Statistical Manifolds}
\label{SS:ApproxDist}

\begin{figure}[t]
  \centerline{
    \includegraphics[scale=.55]{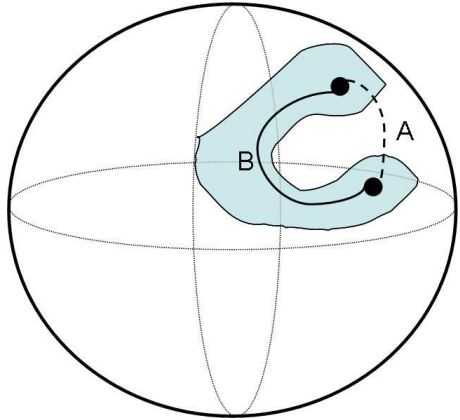}
    }
  \caption{The Fisher information distance between points cannot be exactly calculated about a manifold if the data exists on a submanifold of interest (shaded area). Rather than directly calculating the distance between points (A) , the distance should be approximated by a geodesic along the submanifold (B).}
  \label{f:submanifold}
\end{figure}

We have shown the approximation function $\hat{D}_F(p_1,p_2)$ of the Fisher information distance between $p_1$ and $p_2$ can be calculated using a variety of metrics as $p_1\to p_2$. If $p_1$ and $p_2$ do not lie closely together on the manifold, these approximations become weak. An example of this is illustrated in Fig.~\ref{f:submanifold}, where the manifold of interest lies in a subspace of another manifold, and the distance between two points should be considered as the distance traveled on the manifold of interest. A good approximation can still be achieved if the manifold is densely sampled between the two end points. By defining the path between $p_1$ and $p_2$ as a series of connected segments and summing the length of those segments, we approximate the distance of the \emph{geodesic}, which is the shortest path along the manifold. Specifically, given the set of $n$ PDFs parameterized by $\mP_\theta=\left\{\theta_1,\ldots,\theta_n\right\}$, the Fisher information distance between $p_1$ and $p_2$ can be estimated as:

\[
D_F(p_1,p_2)\approx\min_{m,\left\{\theta_{(1)},\ldots,\theta_{(m)}\right\}} {\sum_{i=1}^m{ D_F(p(\theta_{(i)}),p(\theta_{(i+1)}))}},\quad p(\theta_{(i)})\rightarrow p(\theta_{(i+1)})\: \forall \: i
\]
where $p(\theta_{(1)})=p_1$, $p(\theta_{(m)})=p_2$, $\left\{\theta_{(1)},\ldots,\theta_{(m)} \right\} \in \mP_\theta$, and $m\leq n$.

\begin{figure}[t]
  \centerline{
    \includegraphics[scale=.55]{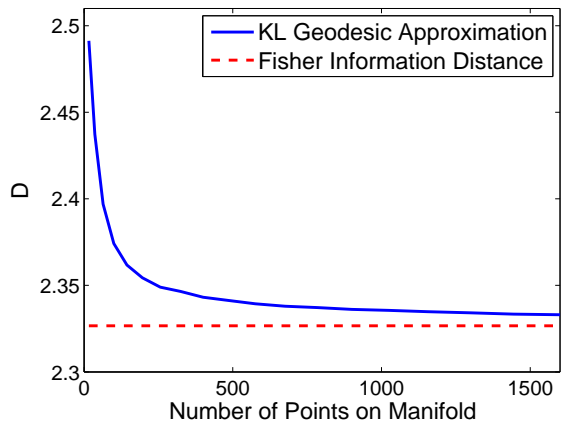}
    }
  \caption{Convergence of the graph approximation of the Fisher information distance using the Kullback-Leibler divergence. As the manifold is more densely sampled, the approximation approaches the true value.}
  \label{f:kl_converge}
\end{figure}

Using our approximation of the Fisher information distance as $p_1\to p_2$ (whether KL-divergence or Hellinger distance is of no immediate concern), we can now define an approximation function $G$ for all pairs of PDFs:
\Eq{ \label{Eq:FID_Approx_G}
G(p_1,p_2;\mP)=\min_{m,\mP} {\sum_{i=1}^m{ \hat{D}_F(p_{(i)},p_{(i+1)})}}, \quad p_{(i)}\rightarrow p_{(i+1)}\: \forall \: i}
where $\mP=\left\{p_1,\ldots,p_n\right\}$ is the available collection of PDFs on the manifold. Intuitively, this estimate calculates the length of the shortest path between points in a connected graph on the well sampled manifold, and as such $G(p_1,p_2;\mP)\to D_F(p_1,p_2)$ as $n\to\infty$. This is similar to the manner in which Isomap \cite{Tenenbaum&etal:Science00} approximates distances on Euclidean manifolds. Figure \ref{f:kl_converge} illustrates this approximation by comparing the KL graph approximation to the actual Fisher information distance for the univariate Gaussian case. As the manifold is more densely sampled (uniformly in mean and variance parameters for this simulation), the approximation converges to the true Fisher information distance, as calculated in (\ref{Equation:FID_normal}).

\subsection{Dimensionality Reduction}
\label{SS:DimRed}
Given a matrix of dissimilarities between entities, many algorithms have been developed to find a low-dimensional embedding of the original data $\psi:\mM\rightarrow \Real^d$. These techniques have been classified as a group of methods called Multi-Dimensional Scaling (MDS). There are supervised methods, which are generally used for classification purposes, and unsupervised methods, which are often used for clustering and manifold learning. Using these MDS methods allows us to find a single low-dimensional coordinate representation of each high-dimensional, large sample, data set.

\subsubsection{Classical Multi-Dimensional Scaling}
\label{SS:cMDS}

Classical MDS (cMDS) takes a matrix of dissimilarities and embeds each point into a Euclidean space. This is performed by first centering the dissimilarities about the origin, then calculating the eigenvalue decomposition of the centered matrix. This unsupervised method permits the calculation of the low-dimensional embedding coordinates which reveal any natural separation or clustering of the data.

Define $D$ as a dissimilarity matrix which contains (or approximates) Euclidean distances. Let $B$ be the ``double centered'' matrix which is calculated by taking the matrix $D$, subtracting its row and column means, then adding back the grand mean and multiplying by $-\frac{1}{2}$. As a result, $B$ is a version of $D$ centered about the origin. Mathematically, this process is solved by
\[
B=-\frac{1}{2}H D^2 H
,\]
where $H=I-(1/N)1 1^T$, $I$ is the $N$-dimensional identity matrix, and $1$ is an $N$-element vector of ones.

The embedding coordinates, $\vY\in\Real^{d\times n}$, can then be determined by taking the eigenvalue decomposition of $B$,
\[B=[V_1 V_2] \textrm{diag}\left(\lambda_1,...,\lambda_N\right)[V_1 V_2]^T,\]
and calculating
\[\vY=\textrm{diag}\left(\lambda_1^{1/2},...,\lambda_d^{1/2}\right)V_1^T.\]
The matrix $V_1$ consists of the eigenvectors corresponding to the $d$ largest eigenvalues $\lambda_1,\ldots,\lambda_d$ while the remaining $N-d$ eigenvectors are represented as $V_2$. The term `$\textrm{diag}(\lambda_1,\ldots,\lambda_N)$' refers to an $N\times N$ diagonal matrix with $\lambda_i$ as its $i^{\textrm{th}}$ diagonal element.

\begin{figure}[t]
  \centerline{
  \subfigure[Fisher Information] {\label{f:cmds_fi_normal}
    \includegraphics[scale=.55]{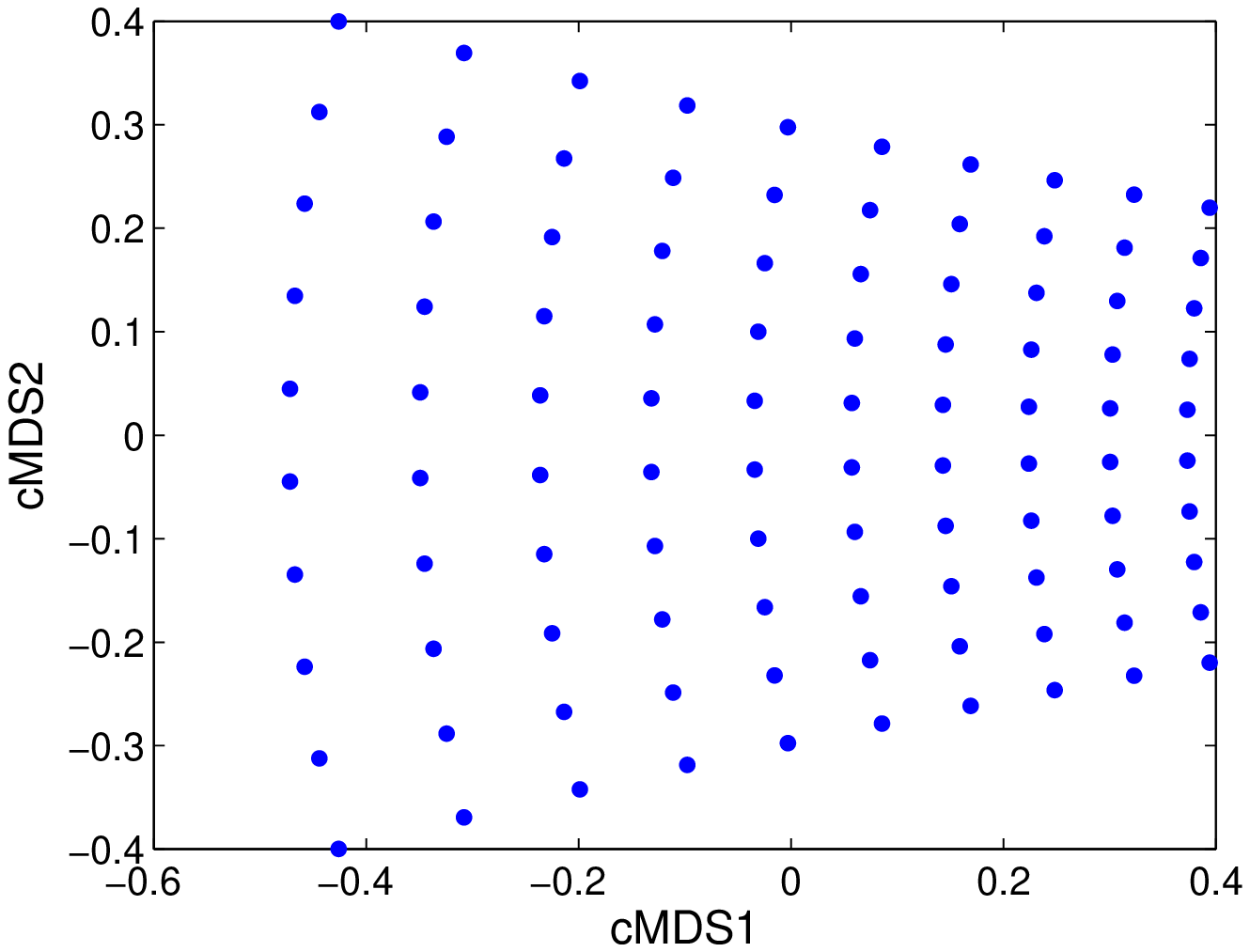}
  }
  \subfigure[Kullback-Leibler Approximation] {\label{f:cmds_kl_normal}
    \includegraphics[scale=.55]{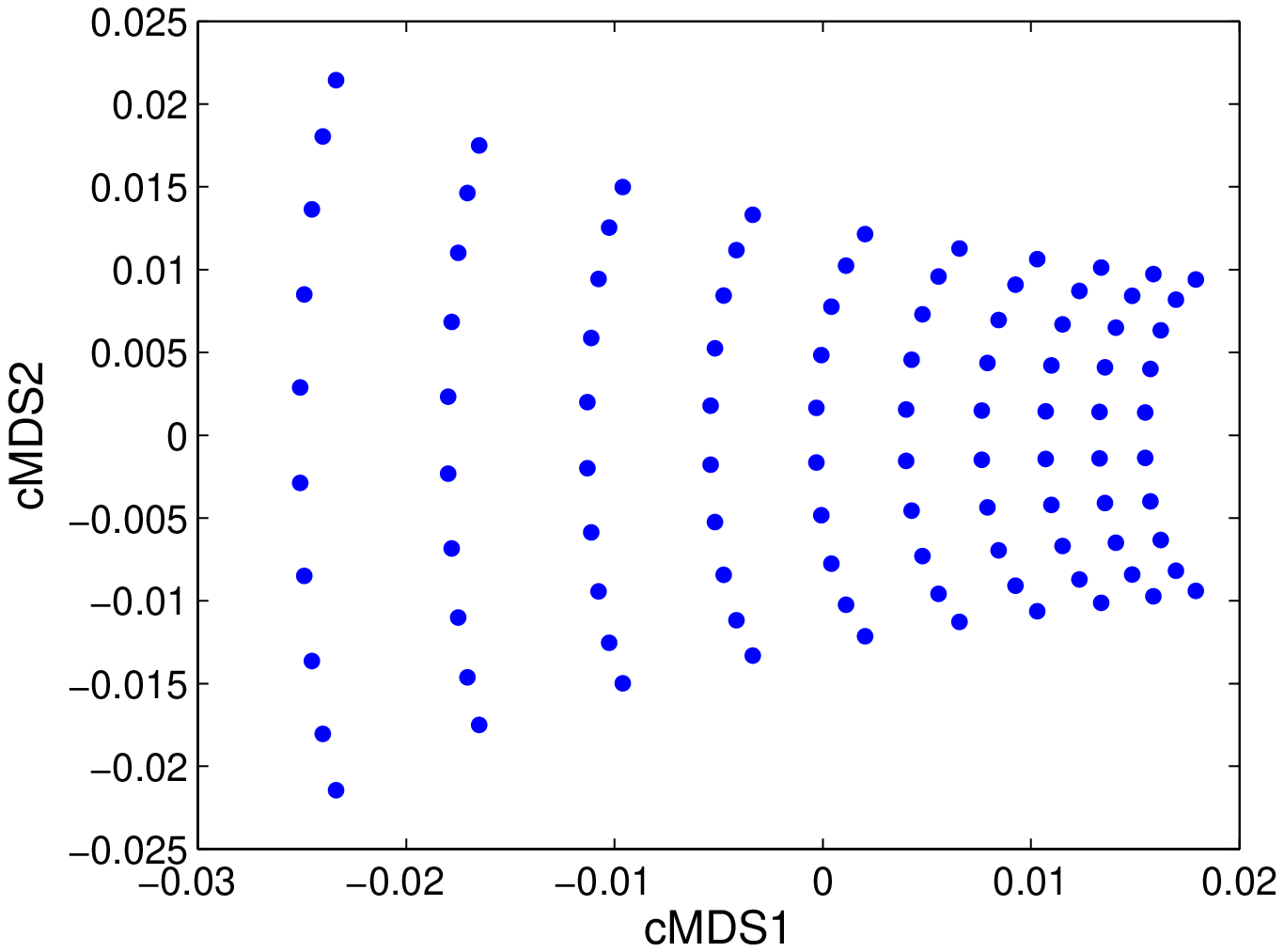}}
  }
  \caption{Classical MDS to the matrix of a) Fisher information distances and b) Kullback-Leibler geodesic approximations of the Fisher information distance, on a grid of univariate normal densities, parameterized by $\left(\mu,\sigma\right)$}
  \label{f:cmds_isomap_fi}
\end{figure}

To continue our illustration from Section \ref{SS:example}, let $D$ be the matrix of Fisher information distances defined in (\ref{Equation:FID_normal}) for the set of univariate normal densities $\mP$, where $D(i,j)=D_F(p_i,p_j)$. Figure \ref{f:cmds_fi_normal} displays the results of applying cMDS to $D$. We demonstrate the embedding with the geodesic approximation of the Fisher information distance (\ref{Eq:FID_Approx_G}) in Fig.~\ref{f:cmds_kl_normal}, which is very similar to the embedding created with the exact values. It is clear that while the densities defining the set $\mP$ are parameterized on a rectangular grid, the manifold on which $\mP$ lives is not rectangular itself, which is due to the differing effects that changes in mean and variance have on the Gaussian PDF.

\subsubsection{Laplacian Eigenmaps}
Laplacian Eigenmaps (LEM) is an unsupervised technique developed by Belkin and Niyogi and first presented in \cite{Belkin&Niyogi:NIPS02}. This performs non-linear dimensionality reduction by performing an eigenvalue decomposition on the graph Laplacian formed by the data. As such, this algorithm is able to discern low-dimensional structure in high-dimensional spaces that were previously indiscernible with methods such as principal components analysis (PCA) and classical MDS. The algorithm contains three steps and works as follows:

\begin{enumerate}
  \item Construct adjacency graph\\
    Given dissimilarity matrix $D_X$ between data points in the set $\vX$, define the graph $G$ over all data points by adding an edge between points $i$ and $j$ if $\vX_i$ is one of the $k$-nearest neighbors of $\vX_j$.
  \item Compute weight matrix $W$\\
    If points $i$ and $j$ are connected, assign $W_{ij}=e^{-\frac{D_X(i,j)^2}{t}}$, otherwise $W_{ij}=0$.
  \item Construct low-dimensional embedding\\
    Solve the generalized eigenvalue problem \[L\textbf{f}=\lambda D\textbf{f},\] where $D$ is the diagonal weight matrix in which $D_{ii}=\sum_j{W_{ji}}$, and $L=D-W$ is the Laplacian matrix. If $\left[\textbf{f}_1,\ldots,\textbf{f}_d\right]$ is the collection of eigenvectors associated with $d$ smallest generalized eigenvalues which solve the above, the $d$-dimensional embedding is defined by $\textbf{y}_i=\left(v_{i1},\ldots,v_{id}\right)^T, 1\le i\le n$.
\end{enumerate}

\subsubsection{Additional MDS Methods}
\label{SS:OtherMDS}
While we choose to only detail the cMDS and LEM algorithms, there are many other methods for performing dimensionality reduction in a linear fashion (PCA) and non-linearly (Local Linear Embedding \cite{Roweis&Saul:Science00}) for unsupervised learning. For supervised learning there are also linear (Linear Discriminant Analysis) and non-linear (Classification Constrained Dimensionality Reduction \cite{Raich&Hero:ICASSP06}, Neighbourhood Component Analysis \cite{Goldberg&Roweis:NIPS04}) methods, all of which can be applied to our framework. We do not highlight the heavily utilized Isomap \cite{Tenenbaum&etal:Science00} algorithm since it is identical to using cMDS on the approximation of the geodesic distances.

\section{Our Techniques}
\label{S:Techniques}
We have presented a series of methods for manifold learning developed in the field of information geometry. By performing dimensionality reduction on a family of data sets, we are able to both better visualize and classify the data. In order to obtain a lower dimensional embedding, we calculate a dissimilarity metric between data sets within the family by approximating the Fisher information distance between their corresponding PDFs. This has been illustrated with the family of univariate normal probability distributions.

In problems of practical interest, however, the parameterization of the probability densities are usually unknown. We instead are given a family of data sets $\mX=\{\vX_1,\vX_2,\ldots,\vX_n\}$, in which we may assume that each data set $\vX_i$ is a realization of some underlying probability distribution to which we do not have knowledge of the parameters. As such, we rely on non-parametric techniques to estimate both the probability density and the approximation of the Fisher information distance. Following these approximations, we are able to perform the same multi-dimensional scaling operations as previously described.

\subsection{Kernel Density Estimation}
\label{SS:KDE}
Kernel methods are non-parametric techniques used for estimating probability densities of data sets. These methods are similar to mixture-models in that they are defined by the normalized sum of multiple densities. Unlike mixture models, however, kernel methods are non-parametric and are comprised of the normalized sum of identical densities centered about each data point within the set (\ref{Equation:KDE}). This yields a density estimate for the entire set in that highly probable regions will have more samples, and the sum of the kernels in those areas will be large, corresponding to a high probability in the resultant density. The kernel density estimate (KDE) of a PDF is defined as

\Eq{ \label{Equation:KDE}
        \hat{p}(x)=\frac{1}{Nh}\sum_{i=1}^N{K\left(\frac{x-x_i}{h}\right)}
    ,}
where $K$ is some kernel satisfying the properties
\[
K(x)\ge0, \,\forall x\in\mX
,\]
\[\int K(x)\,dx=1,\]
and $h$ is the bandwidth or smoothing parameter.

There are two key points to note when using kernel density estimators. First, it is necessary to determine which distribution to use as the kernel. Without a priori knowledge of the original distribution, we choose to use Gaussian kernels,
\Eq{ \label{Equation:GaussianKernel}
        K(\vx)=\frac{1}{(2\pi)^{(d/2)}|\Sigma|^{1/2}}\exp\left(-\frac{1}{2}\vx^T\Sigma^{-1}\vx\right)
    ,}
where $d$ is the dimension of $\vx$ and $\Sigma$ is the covariance matrix, as they have the quadratic properties that will be useful in implementation. Secondly, the bandwidth parameter is very important to the overall density estimate. Choosing a bandwidth parameter too small will yield a peak filled density, while a bandwidth that is too large will generate a density estimate that is too smooth and loses most of the features of the distribution. There has been much research done in calculating optimal bandwidth parameters, resulting in many different methods \cite{Silverman:86,Terrell:JASA90} which can be used in our framework.

We note that the mean squared error of a KDE decreases only as $n^{-\rm{O}(1/d)}$, which becomes extremely slow for large $d$. As such, it may be difficult to calculate good kernel density estimates. However, for our purposes, the estimation of densities is secondary to the estimation of the divergence between them. As such, the issues with MSE of density estimates in large dimensions, while an area for future work, is not of immediate concern.

\subsection{Algorithm}
\label{SS:Algo}
\begin{algorithm}[t]
\caption{Fisher Information Non-parametric Embedding}
\label{algo:full}
    \begin{algorithmic}[1]
        \REQUIRE Collection of data sets $\mX=\{\vX_1,\vX_2,\ldots,\vX_N\}$; the desired embedding dimension $d$
        \FOR{$i=1$ to $N$}
            \STATE Calculate $\hat{p}_i(\vx)$, the density estimate of $\vX_i$
        \ENDFOR
        \STATE Calculate $G$, where $G(i,j)=\hat{D}_F(p_i,p_j)$, the geodesic approximation of the Fisher information distance
        \STATE $\vY=\textrm{embed}(G,d)$ \label{test}
        \ENSURE $d$-dimensional embedding of $\mX$, into Euclidean space $\vY\in \Real^{d\times N}$
    \end{algorithmic}
\end{algorithm}

Fisher Information Non-parametric Embedding (FINE) is presented in Algorithm~\ref{algo:full} and combines all of the methods we have presented in order to find a low-dimensional embedding of a collection of data sets. If we assume each data set is a realization of an underlying PDF, and each of those distributions lie on a manifold with some natural parameterization, then this embedding can be viewed as an embedding of the actual manifold into Euclidean space. Note that in line \ref{test}, `embed$(G,d)$' refers to using any multi-dimensional scaling method (such as cMDS, Laplacian Eigenmaps, etc.) to embed the dissimilarity matrix $G$ into a Euclidean space with dimension $d$.

\section{Applications}
\label{S:Apps}
We have illustrated the uses of the presented framework in the previous sections with a manifold consisting of the set of univariate normal densities, $\mP$. We now present several synthetic and practical applications for the framework, all of which are based around visualization and classification. In each application, the densities are unknown, but we assume they lie on a manifold with some natural parameterization.

\subsection{Simulated Data}
\begin{figure}[t]
  \centerline{
  \subfigure[Swiss Roll] {\label{f:swiss_roll}
    \includegraphics[scale=.55]{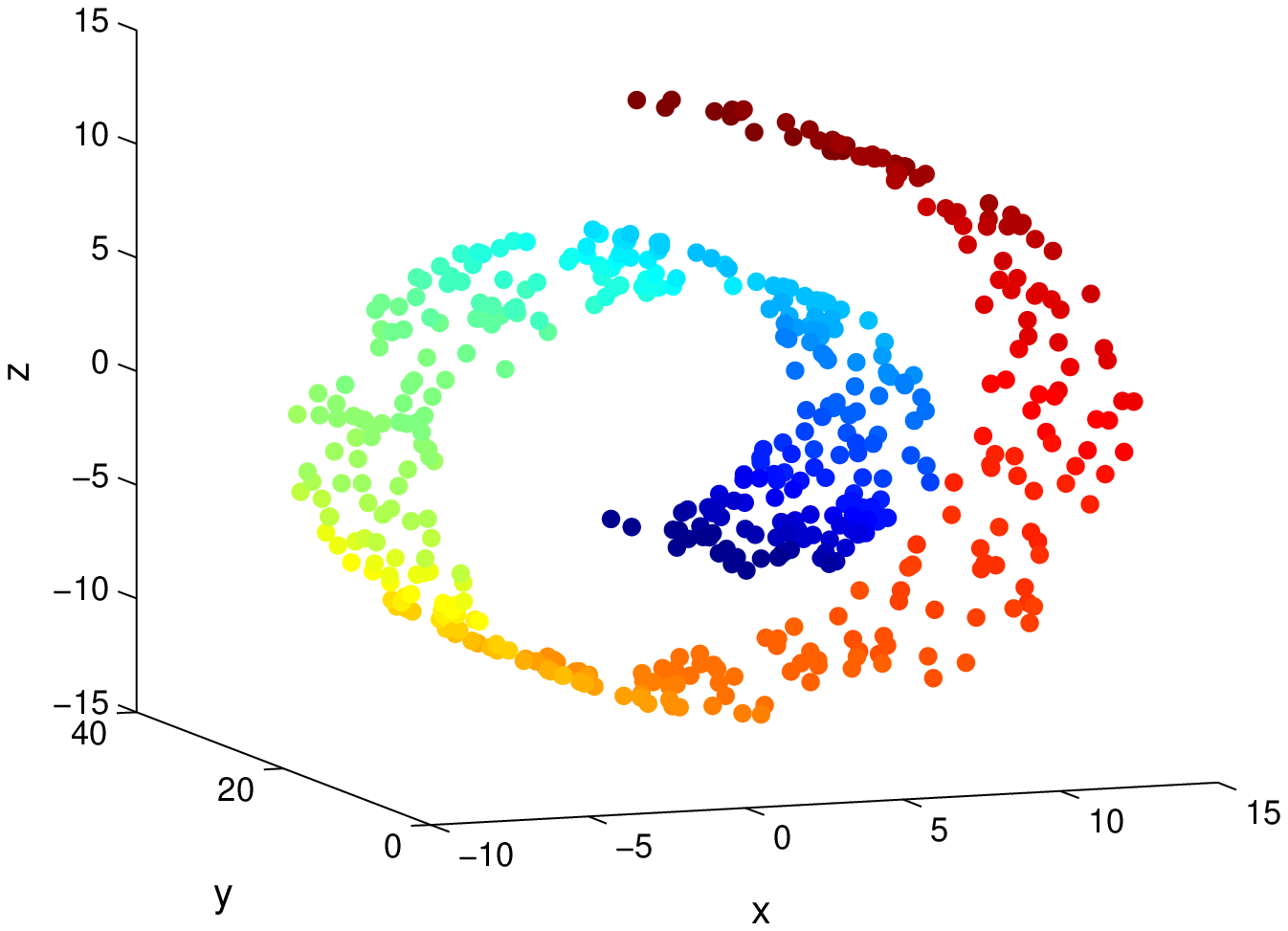}\\
  }
  \subfigure[FINE embedding] {\label{f:swiss_roll_recon}
    \includegraphics[scale=.55]{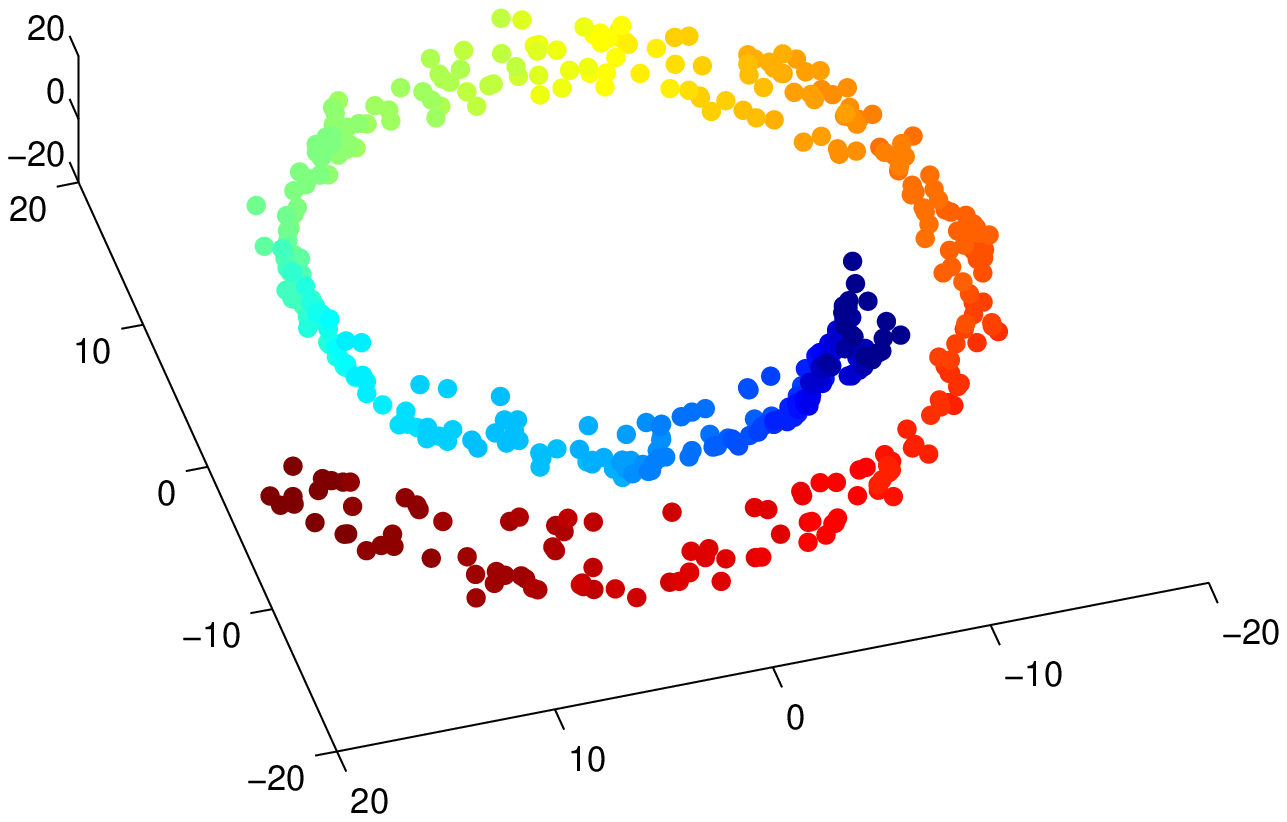}\\}
  }
  \caption{Given a collection of data sets with a Gaussian distribution having means equal to points a sampled `swiss roll' manifold, our methods are able to reconstruct the original statistical manifold from which each data set is derived.}
  \label{f:swiss_roll_ex}
\end{figure}

To demonstrate the ability of our methods to reconstruct the statistical manifold, we create a known manifold of densities. Let $\vY=\{y_1,\ldots,y_n\}$, where each $y_i$ is uniformly sampled on the `swiss roll' manifold (see Fig.~\ref{f:swiss_roll}). Let $\mX=\{\vX_1,\vX_2,\ldots,\vX_n\}$ where each $\vX_i$ is generated from a normal distribution $\mN(y_i,\Sigma)$, where $\Sigma$ is held constant for each density. As such, we have developed a statistical manifold of known parameterization, which is sampled by known PDFs. Utilizing FINE in an unsupervised manner, we are able to recreate the original manifold $\vY$ strictly from the collection of data sets $\mX$. This is shown in Fig.~\ref{f:swiss_roll_recon} where each set is embedded into 3 cMDS dimensions, and the `swiss roll' is reconstructed. While this embedding could easily be constructed using the mean of each set $\vX_i$ as a Euclidean location, it illustrates that FINE can be used for visualizing the statistical manifold as well, without a priori knowledge of the data.

\subsection{Flow Cytometry}
\label{SS:FlowCyt}

\begin{figure}[t]
\center
  \includegraphics[scale=.35]{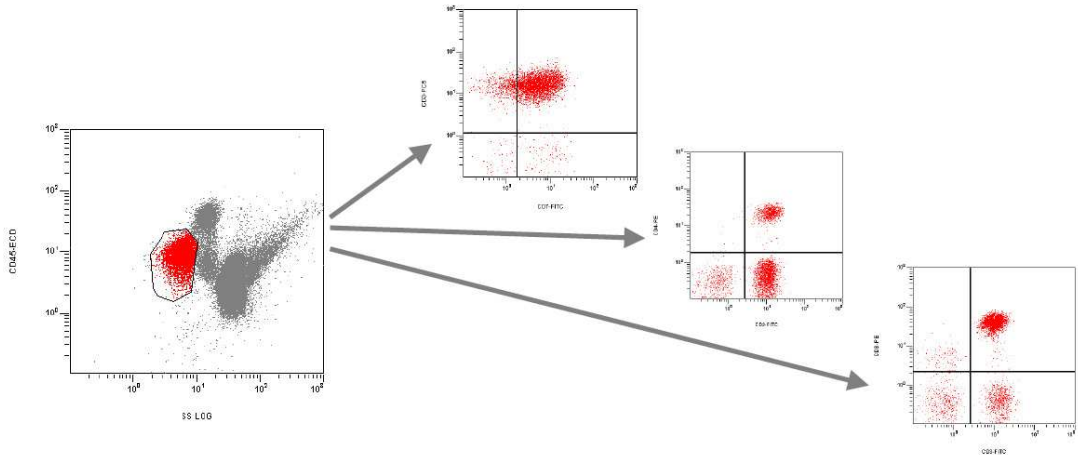}
  \caption{Historically, the process of clinical flow cytometry analysis relies on a series of 2-dimensional scatter plots in which cell populations are selected for further evaluation. This process does not take advantage of the multi-dimensional nature of the problem.}
  \label{f:cytometry_process}
\end{figure}

In clinical flow cytometry, cellular suspensions are prepared from patient samples (blood, bone marrow, and solid tissue), and evaluated simultaneously for the presence of several expressed surface antigens and for characteristic patterns of light scatter as the cells pass through an interrogating laser.
Antibodies to each target antigen are conjugated to fluorescent markers, and each individual cell is evaluated via detection of the fluorescent signal from each marker.  The result is a characteristic multi-dimensional distribution that, depending on the panel of markers selected, may be distinct for a specific disease entity.  The data from clinical flow cytometry can be considered multi-dimensional both from the standpoint of multiple characteristics measured for each cell, and from the standpoint of thousands of cells analyzed per sample.  Nonetheless, clinical pathologists generally interpret clinical flow cytometry results in the form of two-dimensional scatter plots in which the axes each represent one of multiple cell characteristics analyzed (up to 8 parameters per cell in routine clinical flow cytometry, and many more parameters per cell in research applications).  Additional parameters are often utilized to ``gate'' (i.e. select or exclude) specific cell sets based on antigen expression or light scatter characteristics; however, clinical flow cytometry analysis remains a step-by-step process of 2-dimensional histogram analysis (Fig.~\ref{f:cytometry_process}), and the multidimensional nature of flow cytometry is routinely underutilized in clinical practice.

\begin{figure}[t]
  \centerline{
\subfigure[Scatter Plot]{
  \includegraphics[scale=.55]{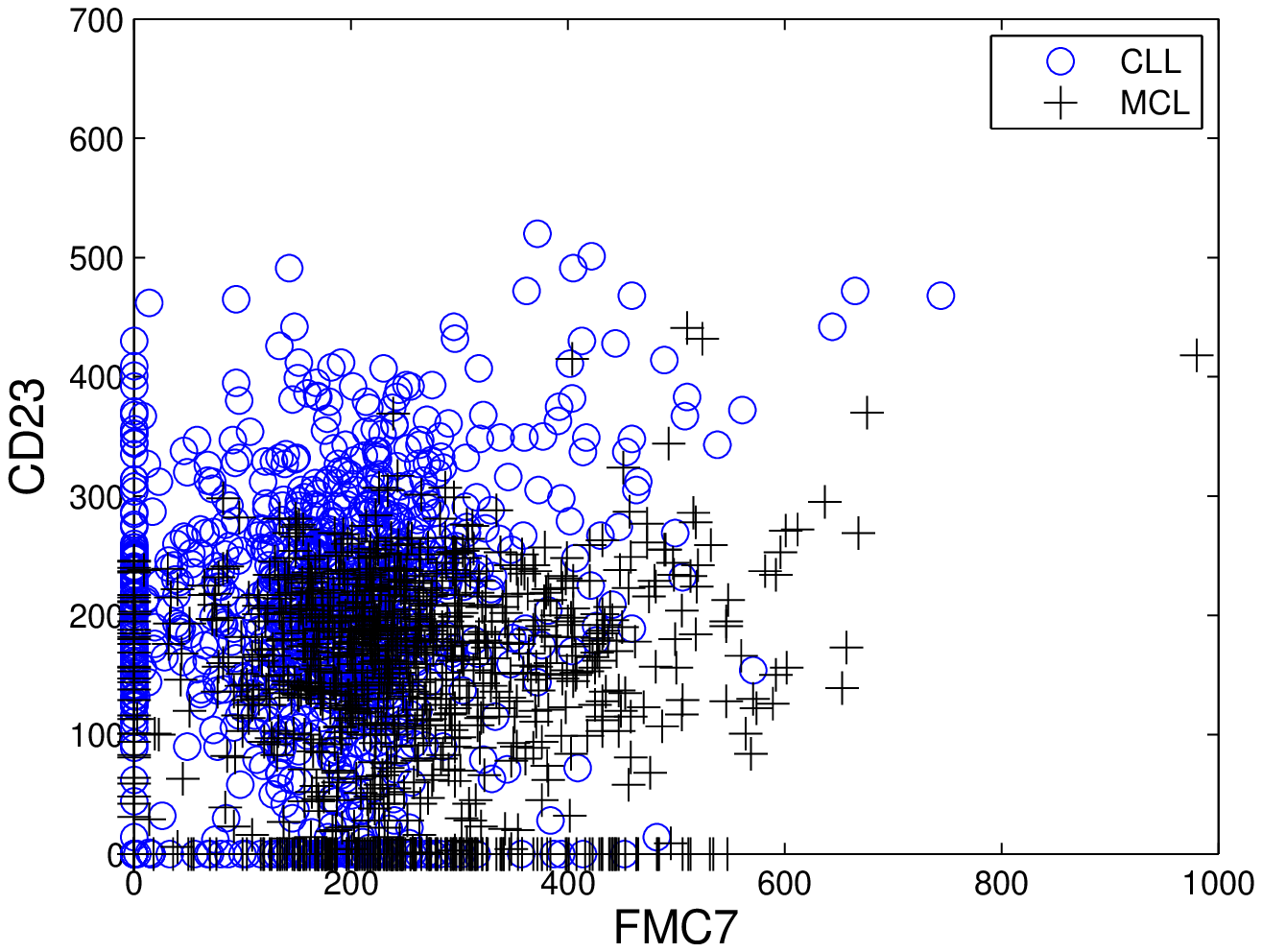}
}
\subfigure[Contour Plot]{
  \includegraphics[scale=.55]{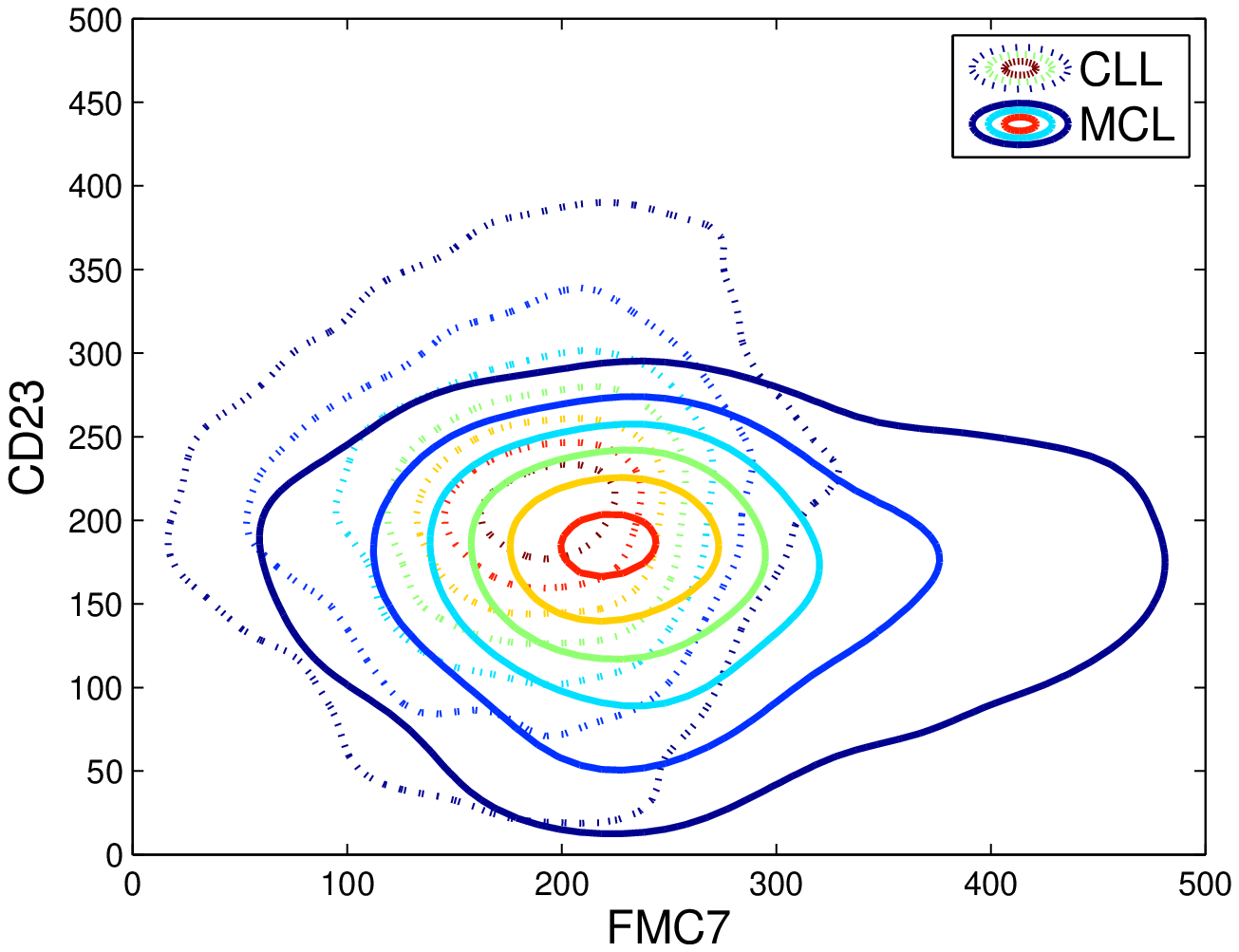}\\
}}
  \caption{2-dimensional plots of disease classes CLL and MCL. The overlapping nature of the scatter plots makes it difficult for pathologists to differentiate disease classes using primitive 2-dimensional axes projections.}
  \label{f:cll_vs_mcl}
\end{figure}

An example of the difficulty in analysis of 2-dimensional scatter plots is illustrated in Fig.~\ref{f:cll_vs_mcl}. Two distinct disease classes, mantle cell lymphoma (MCL) and chronic lymphocytic leukemia (CLL), are illustrated with both scatter and contour plots. Each point represents a distinct blood cell from two different patients, each containing one of the specified diseases; the axes represent those which pathologists have determined to be the two markers which are most differentiating for these two disease classes. It is clear that for these two patients there is significant similarity in the scatter and contour plots of the data. The overlapping nature of these 2-dimensional scatter plot leads to a very primitive analysis of the available data. It would be potentially beneficial, therefore, to develop systems for clustering and classification of clinical flow cytometry data that utilize all dimensions of data derived for each cell during routine clinical analysis.  The variability of distributions of data in multidimensional flow cytometry over various patients is smaller than that associated with a general characterization of a multivariate distribution.  This leads us to believe that these distributions exist on some manifold with a much lower dimensional parameterization.  Hence, we should be able to use FINE for the purpose of viewing a natural clustering of different patients into their respective disease classes based on the full set of markers evaluated in each multiparameter flow cytometric analysis.

For this analysis, we will compare patients with two distinct but immunophenotypically similar forms of lymphoid leukemia - mantle cell lymphoma (MCL) and chronic lymphocytic leukemia (CLL), as illustrated in Fig.~\ref{f:cll_vs_mcl}. These diseases display similar characteristics with respect to many expressed surface antigens, but are generally distinct in their patterns of expression of two common B lymphocyte antigens CD23 and FMC7 (a distinct conformational epitope of the CD20 antigen).  Typically, CLL is positive for expression of CD23 and negative for expression of FMC7, while MCL is positive for expression of FMC7 and negative for expression of CD23.  These distinctions should lead to a difference in densities between patients in each disease class, and should show a natural clustering.

Let $\mX=\{\vX_1,\vX_2,\ldots,\vX_n\}$ where $\vX_i$ is the data set corresponding to the flow cytometer output of the $i^{th}$ patient. Each patient's blood is analyzed for 5 parameters: forward and side light scatter, and 3 fluorescent markers (CD45, CD23, FMC7). Hence, each data set $\vX_i$ is 5-dimensional with $n_i$ elements corresponding to individual blood cells (each $n_i$ may be different). Given that $\mX$ is comprised of both patients with CLL and patients with MCL, we wish to analyze the performance of FINE for the visualization and clustering of cytometry data.

\begin{figure}[t]
  \centerline{
  \includegraphics[scale=.55]{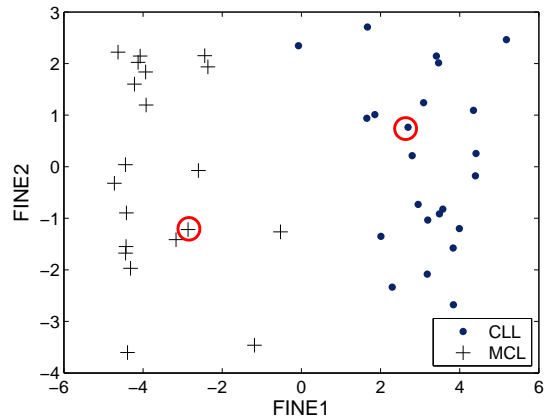}}
  \caption{2-dimensional embedding of CLL ($\bullet$) and MCL ($+$) patients using FINE with cMDS and the Kullback-Leibler divergence as a dissimilarity metric. The circled points correspond to the CLL and MCL cases highlighted in Fig.~\ref{f:cll_vs_mcl}, which are difficult to discern with scatter plots, but well separated in the FINE space.}
  \label{f:fmc7cd23}
\end{figure}

The data set consists of 23 patients with CLL and 20 patients with MCL. The set $\vX_i$ for each patient is on the order of $n_i\approx5000$ cells. The data and clinical diagnosis for each patient was provided by the Department of Pathology at the University of Michigan. Figure \ref{f:fmc7cd23} shows the 2-dimensional embedding with FINE, using cMDS and the Kullback-Leibler divergence set as the dissimilarity metric. Each point in the plot represents an individual patient. Although the discussed methods perform the dimensionality reduction and embedding in unsupervised methods, we display the class labels as a means of analysis. It should be noted that there exists a natural separation between the different classes. As such, we can conclude that there is a natural difference in probability distribution between the disease classes as well. Although this is known through years of clinical experience, we were able to determine this without any a priori knowledge; simply with a density analysis.

An important byproduct of this natural clustering is the ability to visualize the cytometry data in a manner which allows comparisons between patients. The circled points in Fig.~\ref{f:fmc7cd23} correspond to the patients illustrated in Fig.~\ref{f:cll_vs_mcl}, which were difficult to differentiate by using a scatter plot of the most discerning marker combination as deemed by pathologists. In the space defined by FINE, the patients are easily differentiated and lie well within the clusters of each disease type. By using the embedding created with FINE, pathologists are able to determine similarities between patients, which gives them a quick and easy means of determining which data sets may need further investigation (i.e. for possible misdiagnosis).

\subsection{Document Classification}
Recent work has shown in interest in using dimensionality reduction for the purposes of document classification \cite{Kim:JMLR05} and visualization \cite{Huang:CMVEB05}. Typically documents are represented as very high-dimensional PDFs, and learning algorithms suffer from the \emph{curse of dimensionality}. Dimensionality reduction not only alleviates these concerns, but it also reduces the computational complexity of learning algorithms due to the resultant low-dimensional space. As such, the problem of document classification is an interesting application for FINE.

Given a collection of documents of known class, we wish to best classify a document of unknown class. A document can be viewed as a realization of some overriding probability distribution, in which different distributions will create different documents. For example, in a newsgroup about computers you could expect to see multiple instances of the term ``laptop'', while a group discussing recreation may see many occurrences of ``sports''.  The counts of ``laptop'' in the recreation group, or ``sports'' in the computer group would predictably be low. As such, the distributions between articles in computers and recreation should be distinct. In this setting, we defined the PDFs as the \emph{term frequency} representation of each document. Specifically, let $x_i$ be the number of times term $i$ appears in a specific document. The PDF of that document can then be characterized as the multinomial distribution of normalized word counts, with the maximum likelihood estimate provided as
\Eq{ \label{E:tf}
\hat{p}(x)=\left(\frac{x_1}{\sum_i{x_i}},\ldots,\frac{x_N}{\sum_i{x_i}}\right)
.}

By utilizing the term frequencies as a multinomial distribution, and not implementing a kernel density estimator, we show that our methods are not tied to the KDE, but we simply use it in the case of continuous densities as a means of estimation. If one has a priori knowledge of the distribution, that step is unnecessary. Additionally, we use the Hellinger distance due to the multinomial nature of the distribution. As described in Section \ref{SS:Others}, $D_H$ has a monotonic transformation to $D_C$, which is the natural metric on the sphere defined by multinomial PDFs.

For illustration, we will utilize the well known 20 Newsgroups data set\footnote{\url{http://people.csail.mit.edu/jrennie/20 Newsgroups/}}, which is commonly used for testing document classification methods. This set contains word counts for postings on 20 separate newsgroups. We choose to restrict our simulation to the 4 domains with the largest number of sub-domains (comp.*, rec.*, sci.*, and talk.*), and wish to classify each posting by its highest level domain. Specifically we are given $\mP=\left\{p_1,\ldots,p_N\right\}$ where each $p_i$ corresponds to a single newsgroup posting and is estimated with (\ref{E:tf}). We note that the data was preprocessed to remove all words that occur in 5 or less documents\footnote{\url{http://www.cs.uiuc.edu/homes/dengcai2/Data/TextData.html}}.

\subsubsection{Unsupervised FINE}
\begin{figure*}[t]
  \centerline{
  \subfigure[FINE] {\label{f:fine} {
    \includegraphics[scale=.55]{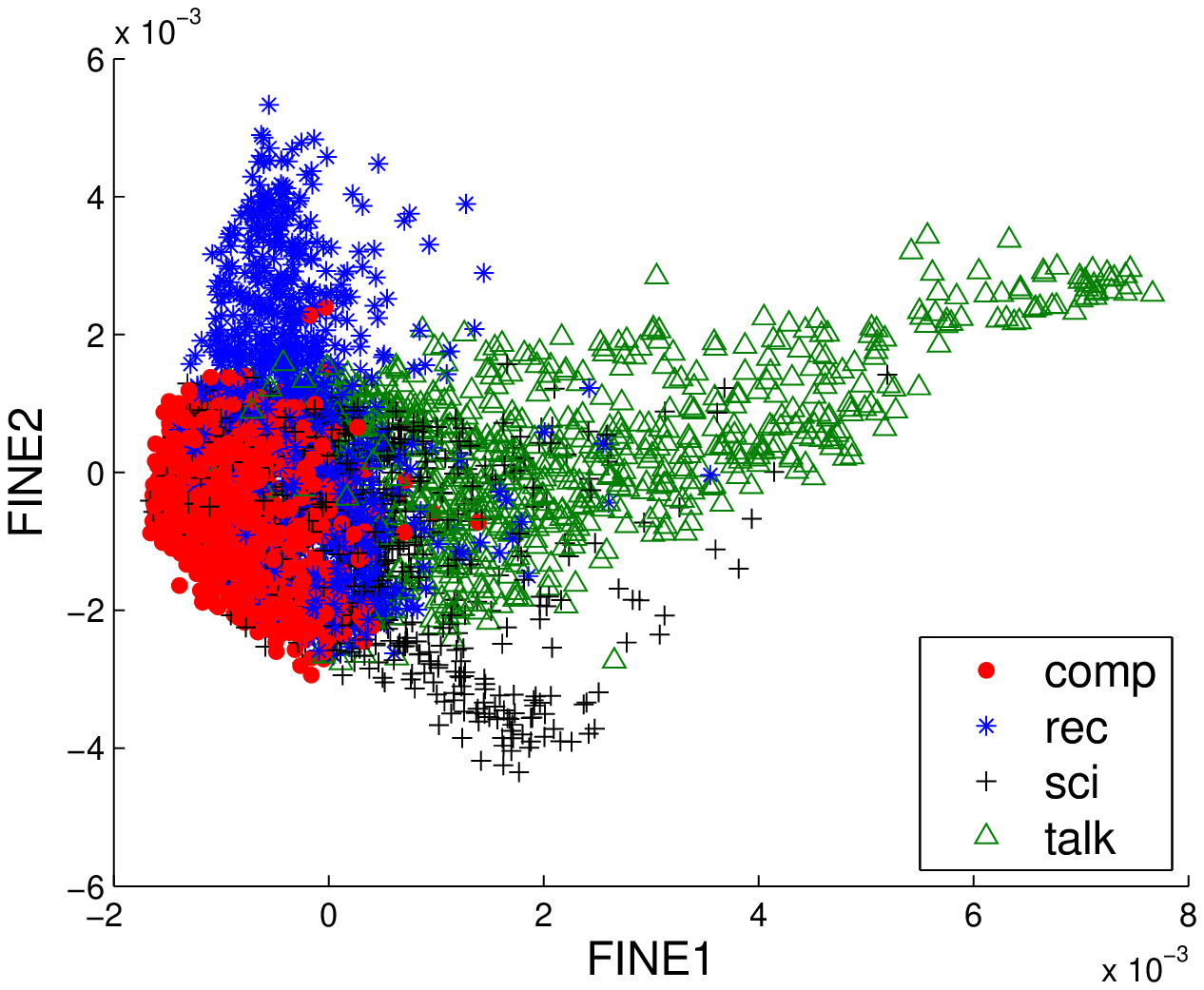}}
  }
  \subfigure[PCA] {\label{f:pca} {
    \includegraphics[scale=.55]{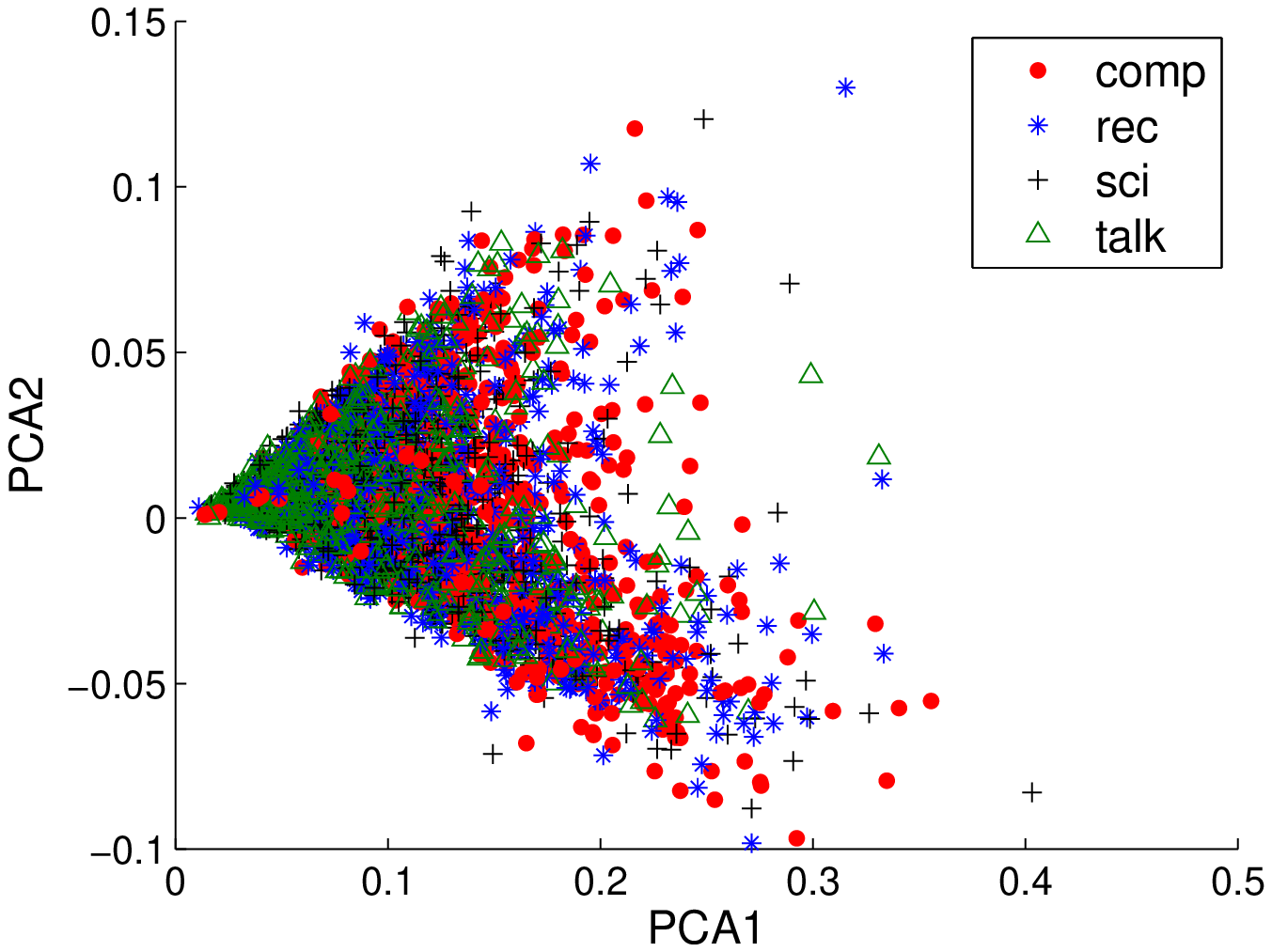}}
  }
  }
  \caption{2-dimensional embeddings of 20 Newsgroups data. The data displays some natural clustering, in the information based embedding, while the PCA embedding does not distinguish between classes.}
  \label{f:20news_4groups_lapeig}
\end{figure*}

First, we utilize unsupervised methods to see if the natural geometry exists between domains. Using Laplacian Eigenmaps on the dissimilarities calculated with the Hellinger distance, we found an embedding $\mP\to\Real^2$. Figure \ref{f:fine} shows the natural geometric separation between the different document classes, although there is some overlap (which is to be expected). Contrarily, a Principal Components Analysis (PCA) embedding (Fig.~\ref{f:pca}) does not demonstrate the same natural clustering. PCA is often used as a means to lower the dimension of data for learning problems due to its optimality for Euclidean data. However, the PCA embedding of the 20 Newsgroups set does not exhibit any natural class separation due to the non-Euclidean nature of the data.

\begin{figure}[t]
  \centerline{
  \includegraphics[scale=.55]{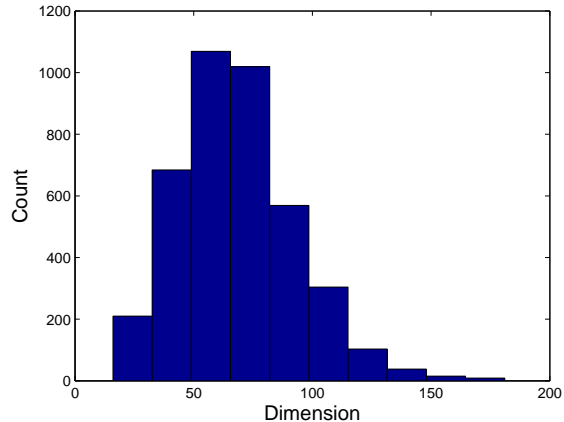}}
  \caption{Local dimension estimates for each document from a random subset of 4020 documents in the 20 Newsgroups data set.}
  \label{f:20hist}
\end{figure}

We now compare the classification performance of FINE to that of PCA. In the case of document classification, dimensionality reduction is important as the natural dimension (i.e. number of words) for the 20 Newsgroups data set is $26,214$. Using local intrinsic dimension estimation \cite{Carter&Hero:SSP07}, Fig.~\ref{f:20hist} shows the histogram of the true dimensionality of the sample documents, so we test performance for low-dimensional embeddings $\mP\to\Real^d$ for $d\in[5,95]$. Following each embedding, we apply an SVM with a linear kernel to classify the data in an `all-vs-all' setting (i.e. classify each test sample as one of 4 different potential classes in a single event, rather than 4 separate binary events). The training and test sets were separated according to the recommended indices, and each set was randomly sub-sampled for computational purposes, keeping the ratio of training to test samples constant (2413 training samples, 1607 test samples). Both the FINE and PCA settings jointly embed the training and test sets.

\begin{figure}[t]
  \centerline{
  \includegraphics[scale=.55]{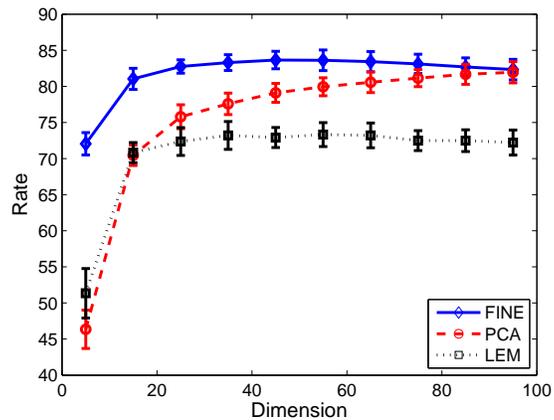}}
  \caption{Classification rates for low-dimensional embedding using different methods for dimensionality reduction. 1-standard deviation confidence intervals shown over 20-fold cross validation.}
  \label{f:info_vs_pca}
\end{figure}

Figure \ref{f:info_vs_pca} illustrates that the embedding calculated with FINE outperforms using PCA as a means of dimensionality reduction. The classification rates are shown with a 1-standard deviation confidence interval, and FINE with a dimension as low as $d=25$ generates results comparable to those of a PCA embedding with $d=95$. To ease any concerns that Laplacian Eigenmaps (LEM) is simply a better method for embedding these multinomial PDFs, we calculated an embedding with LEM in which each PDF was viewed as a Euclidean vector with the $L_2$-distance used as a dissimilarity metric. This form of embedding performed much worse than the information based embedding using the same form of dimensionality reduction and the same linear kernel SVM, while comparable to the PCA embedding in very low dimensions.

\subsubsection{Supervised FINE}
\begin{figure}[t]
  \centerline{
  \includegraphics[scale=.55]{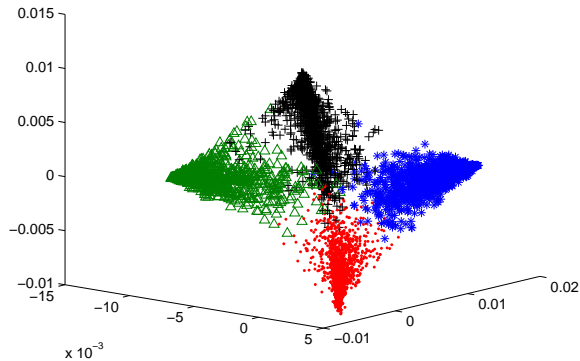}}
  \caption{3-dimensional embedding of 20 Newsgroups corpus using FINE in a supervised manner.}
  \label{f:20news_ccdr}
\end{figure}

If we allow FINE to use supervised methods for embedding, we can dramatically improve classification performance. By embedding with Classification Constrained Dimensionality Reduction (CCDR) \cite{Raich&Hero:ICASSP06}, which is essentially LEM with an additional tuning parameter defining the emphasis on class labels in the embedding, we are able to get good class separation even in 3 dimensions (Fig.~\ref{f:20news_ccdr}). We now compare FINE to the diffusion kernels developed by Lafferty and Lebanon \cite{Lafferty&Lebanon:TIT05} for the purpose of document classification. The diffusion kernels method uses the full term-frequency representation of the data and does not utilize any dimensionality reduction. We stress this difference to determine whether or not using FINE for dimensionality reduction can generate comparable results.

\begin{table}[t!]
\center{
\begin{tabular}{|l|c||r|r||r|r|}
  \multicolumn{2}{c}{} & \multicolumn{2}{c}{FINE} & \multicolumn{2}{c}{Diffusion Kernels} \\ \hline
  Task & $L$ & Mean & STD & Mean & STD \\ \hline
  \multirow{4}{*}{comp.*} & 40 & \textbf{82.3750} & 4.1003 & 75.5750 & 3.9413 \\ 
   & 80 & \textbf{85.8250} & 2.8713 & 83.0250 & 3.4469 \\
   & 120 & \textbf{87.6000} & 2.0876 & 85.5750 & 3.2129 \\
   & 200 & \textbf{87.9750} & 2.3978 & 87.8500 & 2.2775 \\
   & 400 & \textbf{89.8000} & 2.0926 & 89.6250 & 1.9992 \\
   & 600 & 90.6500 & 2.0970 & \textbf{91.3000} & 2.4677 \\
   & 1000 & 91.3000 & 2.3864 & \textbf{91.9000} & 2.2572 \\ \hline\hline
  \multirow{4}{*}{rec.*} & 40 & \textbf{82.3500} & 3.2610 & 76.2000 & 3.1514 \\
   & 80 & \textbf{86.3500} & 2.0462 & 82.0000 & 3.8251 \\
   & 120 & \textbf{87.1500} & 2.3345 & 83.1250 & 3.9599 \\
   & 200 & \textbf{89.5500} & 1.4133 & 86.8750 & 2.1143 \\
   & 400 & \textbf{91.4750} & 2.2152 & 90.7000 & 2.0545 \\
   & 600 & 92.7500 & 1.2722 & \textbf{93.1000} & 2.0494 \\
   & 1000 & 93.2000 & 1.3318 & \textbf{94.6250} & 1.4223 \\ \hline\hline
  \multirow{4}{*}{sci.*} & 40 & \textbf{78.6500} & 2.8102 & 76.3250 & 3.2898 \\
   & 80 & \textbf{80.3750} & 3.3280 & 77.4750 & 4.2286 \\
   & 120 & \textbf{81.5250} & 2.8722 & 78.2250 & 3.1518 \\
   & 200 & \textbf{83.4000} & 2.9585 & 82.2000 & 3.0236 \\
   & 400 & 86.1750 & 2.2021 & \textbf{86.2000} & 2.2325 \\
   & 600 & \textbf{87.1750} & 2.9212 & 87.0500 & 2.9731 \\
   & 1000 & 89.3000 & 2.3022 & \textbf{89.8000} & 2.2384 \\ \hline\hline 
  \multirow{4}{*}{talk.*} & 40 & \textbf{89.1250} & 3.1241 & 82.2750 & 2.9131 \\
   & 80 & \textbf{90.4250} & 2.8895 & 85.9250 & 3.6859 \\
   & 120 & \textbf{91.1250} & 2.5745 & 86.5500 & 4.0161 \\
   & 200 & \textbf{92.6500} & 1.8503 & 89.7750 & 3.1518 \\
   & 400 & \textbf{93.1000} & 1.9775 & 92.4750 & 2.1672 \\
   & 600 & \textbf{94.7500} & 1.3908 & 94.3750 & 1.5634 \\
   & 1000 & \textbf{94.8500} & 1.5483 & \textbf{94.8500} & 1.4244 \\
  \hline
\end{tabular}}
\caption{Experimental results on 20 Newsgroups corpus, comparing FINE using CCDR and a linear SVM to a multinomial diffusion kernel based SVM. The performance (classification rate in \%) is reported as mean and standard deviation for different training set sizes $L$, over a 20-fold cross validation.}
\label{t:rates}
\end{table}

We first illustrate the classification performance in a `one vs. all' setting, in which all samples from a single class were given a positive label (i.e.~$1$) and all remaining samples were labeled negatively (i.e.~$-1$). In the FINE setting, we first subsampled from the training and test sets, using a test set size of $200$, then used CCDR to embed the entire data set into $\Real^{d}$, with $d\in[5,95]$ chosen to maximize classification performance. The classification task was performed using a simple linear kernel SVM, \[K(X,Y)=X\cdot Y.\] For the diffusion kernels setting, \[K(X,Y)=(4\pi t)^{\frac{n}{2}} \exp\left(-\frac{1}{t} \arccos^2 \left(\sqrt{X} \cdot \sqrt{Y}\right)\right),\] we chose parameter value $t$ which optimized the classification performance at each iteration. The experimental results of performance versus training set size, with 20-fold cross validation, are shown in Table \ref{t:rates}, where the highest performance at each range is highlighted. FINE shows a significant performance increase over the diffusion kernels method for sets with low sample size. As the sample size increases, however, the gap in performance between the diffusion kernels method and FINE decreases, with diffusion kernels eventually surpassing FINE. 

\begin{figure}[t]
  \centerline{
  \includegraphics[scale=.55]{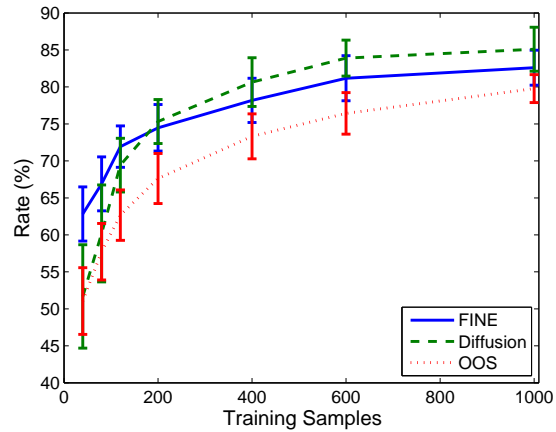}}
  \caption{Classification rates for low-dimensional embedding with FINE using CCDR vs Diffusion kernels. The classification task was all v.s. all. Rates are plotted versus number of training samples. Confidence intervals are shown at one standard deviation. For comparison to the joint embedding (FINE), we also plot the performance of FINE using out of sample extension (OOS).}
  \label{f:fine_ccdr}
\end{figure}

We now modify the classification task from a `one v.s. all' to an `all v.s. all' setting, in which each class is given a different label and the task is to assign each test sample to a specific class. Classification rates are defined as the number of correctly classified test samples divided by the total number of test samples (kept constant at $200$). The structure of the experiment is otherwise identical to the `one v.s. all' setting. We once again notice in Fig.~\ref{f:fine_ccdr} that FINE outperforms the diffusion kernels method for low sample sizes. The point at which the diffusion kernels method surpasses FINE has decreased (i.e. $L\approx200$ for `all v.s. all' compared to $L\approx600$ for `one v.s. all'), yet FINE is still competitive as the sample size increases.

While our focus when using FINE has been on jointly embedding both the training and test samples (while keeping the test samples unlabeled), Fig.~\ref{f:fine_ccdr} also illustrates the use of out of sample extension (OOS) \cite{Raich&Costa:TSP08} with FINE. In this scenario, the training samples are embedded as normal with CCDR, while the test samples are embedded into the low-dimensional space using interpolation. This setting allows for a significant decrease in computational complexity given the fact that the FINE embedding has already been determined for the training samples (i.e. new test samples are received). A decrease in performance exists when compared to the jointly embedded FINE, which is reduced as the number of training samples increases.

Analysis of the results in both the `one v.s. all' and `all v.s. all' cases shows that FINE can improve upon the deficiencies of the diffusion kernels method in the low sample size region. By viewing each document as a coarse approximation of the overriding class PDF, it is easy to see that, for low sample sizes, the estimate of the within class PDF generated by the diffusion kernels will be highly variable, which leads to poor performance. By reducing the dimension with FINE, the variance is limited to significantly fewer dimensions, enabling documents within each class to be drawn nearer to one another. While this could also bring the classes closer to each other, the utilization of CCDR ensures class separation. This results in better classification performance than using the entire multinomial distribution. As the number of training samples increases, the effect of dimensionality is reduced, which allows the diffusion kernels to better approximate the multinomial PDF representative of each class. This reduction in variance across all dimensions ensures that a few anomalous documents will not have the same drastic effect as they would in the low sample size region. As such, the performance gain surpasses that of FINE, due to the fact that the \emph{curse of dimensionality} was alleviated elsewhere (i.e.~increase in sample size). We note that while FINE performs slightly worse than diffusion kernels in the large sample size region, it still performs competitively with a leading classification method which utilizes the full dimensional data.

\begin{figure}[t]
  \centerline{
  \includegraphics[scale=.55]{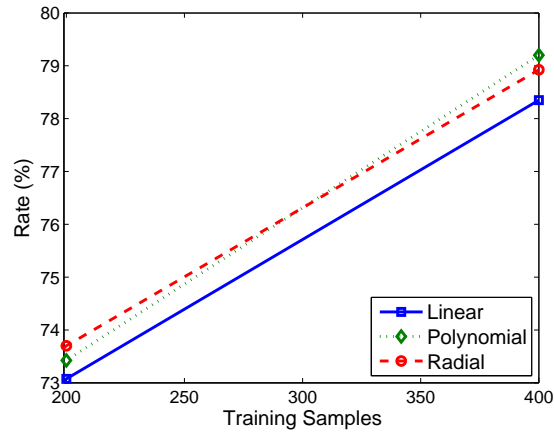}}
  \caption{Comparison of classification performance on the 20 Newsgroups data set with FINE using different SVM kernels; one linear and two non-linear ($\rm2^{nd}$ polynomial and radial basis function).}
  \label{f:nonlin}
\end{figure}

An additional reason for the diffusion kernels improved performance over FINE in the large sample size region is that we have restricted FINE to using a linear kernel for this experiment, while the diffusion kernels method is very non-linear. We do this to show that even a simple linear classifier can perform admirably in the FINE reduced space. Using a non-linear kernel would show increased performance with FINE. This is illustrated in Fig.~\ref{f:nonlin}, where we compare the performance of FINE using an SVM classifier with a linear kernel ($K(X,Y)=X^TY$), $\rm2^{nd}$ degree polynomial kernel ($K(X,Y)=(\gamma X^TY)^2$), and a radial basis function kernel ($K(X,Y)=\exp(-\gamma |X-Y|^2)$), where $\gamma$ is a weighting constant. For visualization purposes, we show the results for only a subset of the training sample range (i.e. $L=[200,400]$), but it is clear that the use of non-linear kernels improves the performance of FINE. The problem of which of the many possible non-linear kernels is optimal remains open and is a subject for future work.



\section{Conclusions}
\label{S:Conclusions}
The assumption that high-dimensional data lies on a Euclidean manifold is based on the ease of implementation due to the wealth of knowledge and methods based on Euclidean space. This assumption is not viable in many problems of practical interest, as there is often no straightforward and meaningful Euclidean representation of the data. In these situations it is more appropriate to assume the data lies on a \emph{statistical} manifold. Using information geometry, we have shown the ability to find a low-dimensional embedding of the manifold, which allows us to not only find the natural separation of the data, but to also reconstruct the original manifold and visualize it in a low-dimensional Euclidean space. This allows the use of many well known learning techniques which work based on the assumption of Euclidean data.

By approximating the Fisher information distance, FINE is able to construct the Euclidean embedding with an information based metric, which is more appropriate for non-Euclidean data. We have illustrated this approximation by finding the length of the geodesic along the manifold, using approximations such as the Kullback-Leibler divergence and the Hellinger distance. The specific metric used to approximate the Fisher information distance is determined by the problem, and FINE is not tied to any specific choice of metric. Additionally, we point out that although we utilize kernel methods to obtain PDFs, the method used for density estimation is only of secondary concern. The primary focus is the measure of dissimilarity between densities, and the method used to calculate those PDFs is similarly determined by the problem.

We have illustrated FINE's ability to be used in a variety of learning tasks such as visualization, clustering, and classification. FINE is a framework that can be used for a multitude of problems which may seem to have little to nothing in common, such as flow cytometry and document classification. The only commonality between the problems is that each are based around data which has no straightforward Euclidean representation, which is the only setting needed to utilize FINE.  In future work we plan to utilize different classification methods (such as $k$-NN and using different SVM kernels) to maximize our document classification performance. This includes constraining our dimensionality reduction to a sphere, which will allow the use of diffusion kernels in a low-dimensional space. We also plan to continue studies on the effect of using out of sample extension on our performance. Lastly, we will continue to find applications which fit the setting for FINE, such as internet anomaly detection and face recognition, and determine whether or not these problems would benefit from our framework.

\section{Special Thanks}
\label{S:Thanks}
We would like to offer a special thanks to the Department of Pathology at the University of Michigan for helping us isolate a problem of strong interest to them, as well as providing a multitude of data for analysis. We would also like to thank Sung Jin Hwang of the University of Michigan for help with the classification analysis and implementation of the SVMs.


\bibliography{ref}
\bibliographystyle{IEEEbib}
\end{document}